\documentclass[final,3p,times]{elsarticle}

\usepackage{amssymb}
\usepackage{amsmath}
\usepackage{hyperref}
\usepackage{multirow}
\usepackage{hyperref}
\usepackage[inkscapeformat=png]{svg}
\usepackage{float}
\usepackage{algorithm}
\usepackage{algpseudocode}
\usepackage{hyperref} 
\usepackage{amssymb}
\usepackage{pifont}
\usepackage{booktabs}
\newcommand{\cmark}{\ding{51}}%
\newcommand{\xmark}{\ding{55}}%
\journal{Information Fusion}
\newcommand{\hl}[1]{\textcolor{black}{#1}}
\begin{document}

\begin{frontmatter}

\title{Transfer Learning with Foundational Models for Time Series Forecasting using Low-Rank Adaptations}

\author[label1]{M. Germ\'an-Morales}
\author[label1]{A.J. Rivera-Rivas}
\author[label1]{M.J. del Jesus Díaz}
\author[label1,label2]{C.J. Carmona} 

\affiliation[label1]{
            organization={Andalusian Research Institute in Data Science and Computational Intelligence, Department of Computer Science, University of Jaen},
            postcode={E-23071}, 
            state={Jaen},
            country={Spain}}
\affiliation[label2]{
            organization={Leicester School of Pharmacy, DeMontfort University},
            postcode={LE1 7RH},
            state={Leicester},
            country={United Kingdom}}

\begin{abstract}
\hl{Foundational Models are an emerging widely used technique of Generative Artificial Intelligence. These models are distinguished by their scalability and the ease with which they can be adapted through the exploitation of Transfer Learning. The availability of high computational power and large datasets have supported their development, achieving a high generalization capacity due to the enormous and heterogeneous amounts of data used in their initial training. These characteristics contribute to a} solid base that can be adapted or adjusted to a wide range of tasks, increasing their applicability. \hl{This study proposes the methodology LLIAM, a straightforward adaptation of a kind of Foundational Models, Large Language Models, for the Time Series Forecasting task.} \hl{An adequate time-series prompting schema and }Low-Rank Adaptations are used to enhance the knowledge of the model with diverse time series datasets, known as the fine-tuning phase. \hl{A study divided in two stages has been performed for evaluating the effectiveness of the proposed methodology. Initially, a comparison was made between the performance of LLIAM and different state-of-the-art Deep Learning algorithms, including Recurrent Neural Networks and Temporal Convolutional Networks, as well as a LLM-based method, TimeLLM. Following this, a zero-shot study is presented in order to evaluate the generalization capacity of the proposed methodology with time series datasets from unknown domains not considered in the model training. The outcomes of this investigation demonstrate the efficacy of LLIAM, highlighting that this straightforward and general approach can attain competent results without the necessity for applying complex modifications. This work also encourages the use of available resources (such as these pre-trained models) \hl{and efficient fine-tuning techniques} to avoid unnecessary and costly training, narrowing the gap between the goals of traditional Artificial Intelligence and Green Artificial Intelligence.}
\end{abstract}

\begin{keyword}
Time Series Forecasting \sep Transfer Learning \sep Foundational Models \sep Large Language Models \sep Low-Rank Adaptations



\end{keyword}

\end{frontmatter}

\section{Introduction}
The advent of Foundation Models (FMs) \hl{has set a milestone} in several areas within the Machine Learning (ML) paradigm, and in particular Deep Learning (DL). FMs are a type of models that are trained on a large amount of data from different sources, which they use to \hl{reach} a high generalization capacity \hl{integrating all the knowledge extracted form these sources into them}. Several fields such as Natural Language Processing (NLP) or Computer Vision (CV) \cite{zhou2023comprehensivesurveypretrainedfoundation} have experienced a great evolution due to the good performance of these models in tasks related to the generation of new and meaningful content from training data, known as Generative AI \cite{feuerriegel2024generative}. In addition, their large generalization capacities allow them to be adapted easily and quickly to tasks from unknown domains \hl{through Transfer Learning} \cite{Weiss2016}.

A time-series is a type of data consisting of a set of measurements ordered over time. This implies the existence of temporal dependencies between the different moments observed. Their analysis makes it possible to extract these relationships in order to estimate its future value at any time, which is called Time-Series Forecasting (TSF). The similarity between processing a numerical and a textual series allows intuiting that using FMs dedicated to Language Modelling, such as Large Language Models (LLMs), can be adapted to TSF. Nevertheless, several proposals have argued and demonstrated that LLMs have the potential to perform diverse Time-Series Analysis tasks \cite{jin2024positionlargelanguagemodels}, including the capability of being adequate predictors without supplying additional time-series specific information to them \cite{gruver2023large}.

Due to the great interest in FMs\hl{, especially LLMs,} and the opportunities they offer in the field of TSF, we \hl{were interested in whether: i) can we take advantage of the similarities between text and time-series and the high abstraction capacity of LLMs for the time series task? ii) is a large adaptation of these models necessary to achieve good results? It is for these reasons that we} present a \hl{methodology} that takes advantage of Transfer Learning and exploits the capabilities of a LLM to perform this task. This proposal is notable for the fine-tuning of a well-known pre-trained FM for language modeling called LLaMA \cite{touvron2023llama} using an efficient technique known as Low-Rank Adaptation (LoRA) \cite{hu2021loralowrankadaptationlarge}. A modification of a prompting technique employed for feeding time-series as textual prompts inside these models is also introduced. The \hl{proposal, based on the} combination of these techniques altogether is called the Llama Lora-Integrated Autoregressive Model (LLIAM).

\hl{An experimental study with two groups of experiments has been designed to demonstrate the capabilities of our proposal respecting different conventional DL methods, and LLM-base approaches: its base model (LLaMA) and a state-of-the-art model, TimeLLM.} In the first one, we evaluate how LLIAM performs after fine-tuning it over several datasets and compare it to other approaches used on TSF. In the second one, we measure \hl{the behaviour of LLIAM} when making predictions on data sets \hl{from unknown domains}, which is called zero-shot forecasting.

The use of \hl{approaches as LLIAM}, which undergo intensive pre-training, can facilitate the development of new sustainable and environmentally friendly models, aligning with the objectives of Green AI \cite{BOLONCANEDO2024128096}. \hl{LLIAM} reduces the need for extensive training to address issues such as hyperparameter optimization or the selection of an ideal architecture.

This paper is structured as follows: Section \ref{sec:rw} describes the work related to our proposal, starting with \hl{an introduction to the TSF task and the main DL methods that have been used along time. Next, a description of Transfer Learning, Pre-trained Model, FM and LLM with proposals that have benefited from their properties applied to TSF is presented.} Section \ref{sec:lliam} presents our proposal, LLIAM, an adaptation of the famous LLaMA efficiently adapted to \hl{the TSF task}. Section \ref{sec:exp-framework} presents the experimental framework, detailing the materials and methods used for the two experiments performed and how each model is evaluated. The results are also presented in this section. Finally, Section \ref{sec:discussion} concludes the article with some remarks drawn from the experiments conducted.

\section{Related Work}
\label{sec:rw}
Throughout the last years, a wide number of methods have been designed to perform time-series analysis tasks, including TSF. In Section \ref{subsec:tp}, \hl{the fundamental concepts for understanding the TSF problem and a brief explanation of the main DL methods that have been used are introduced. After that, the definition of Transfer Learning is discussed in Section \ref{subsec:tl} with special emphasis on the definitions of Pretrained Models, FM and LLM and how their propoerties have caused a paradigm-shift in the resolution of the TSF task. Finally, in Section \ref{subsec:ft-llm} is presented some of the most relevant tuning techniques that allow these models to be adapted to a wide range of tasks.}

\hl{\subsection{Temporal prediction}}
\label{subsec:tp}
In our day-to-day life, there are numerous phenomena that exhibit different behaviors depending on the time at which they are observed. The measurement of the air temperature at a weather station, the occupancy of any type of service or the stock of specific products are examples whose activity is influenced by the presence of strong or weak time dependencies. Following the example of temperature, it is logical that at noon it is higher than in the morning or at night. Moreover, depending on the season, \hl{its variation may be assumed}. This type of data which is dependent on time is called time-series.

A time-series is composed of a sequentially-collected set of observations. Depending on the number of variables recorded, they can be univariate or multivariate. A univariate time-series only collects one variable at a time, while multivariate gather at least two \cite{brockwell2016c8}. Also, if it presents a sampling frequency, it is said to be a discrete time-series. On the other hand, a time-series is considered continuous if it presents observations for every moment in time \cite{brockwell2016c1}.

\hl{The resolution of the TSF task has always been an interesting and challenging one due to the wide range of domains that present time-dependent information. It is not easy to model the temporal dependencies contained in time series data, as how these will develop.} In essence, this task is described as finding a function $f$ capable of generating $h$ predictions $\hat{Y}=\{\hat{y}_{t+1}, \hat{y}_{t+2}, ..., \hat{y}_{t+h}\}$ given $n$ previous observations $S=\{x_0, x_1, x_2, ..., x_n\}$, minimizing the error between them and the real $Y=\{y_{t+1}, y_{t+2}, ..., y_{t+h}\}$ ones. In this context, the number of time-steps to predict is called \textit{forecast horizon} and the historic previous values of a series are called \textit{lags}.

Machine Learning (ML) techniques show considerable potential for solving this task. ML is a subset of Artificial Intelligence that facilitates the capacity for computers to learn from experience and automatically extract patterns from data. However, it should be noted that the implementation of these techniques necessitates a preliminary investment in the curation and enhancement of the quality of the features presented in the data, including the creation of new ones. A subset of ML, called Deep Learning (DL), aims to minimize this effort by integrating the learning of representation into them, stacking several processing or hidden layers responsible for generating representations of the data with different abstraction levels.

Artificial Neural Networks (ANN) are the base models employed in DL, with the artificial neuron constituting the base element of their layers. ANNs are considered universal approximators \cite{Nielsen2018C4}. However, it should be noted that not all ANNs are considered to be deep (Deep Neural Networks, DNNs). \hl{Simple} ANNs with \hl{one hidden layer} are usually considered Shallow Neural Networks (SNNs) \cite{prince2023understanding}.

The great performance of DL in various fields \cite{Sarker2021} resulted in numerous proposals capable of modeling time-series without investing much time performing feature-engineering over them. Sequence-to-sequence (\textit{seq-to-seq}) architectures are usually chosen for tasks in charge of generating a sequence through another sequence \cite{Neubig17}. Encoder-decoder one is the standard used. It employs an encoder ($E$) for generating a latent representation (or context vector, $z$) of the input sequence, summarizing important information extracted from the series into a set of abstract features. This representation is then used by a decoder ($D$) module for generating the desired output. $E$ and $D$ are usually two DL architectures \cite{Vaswani2017attention}. Nevertheless, the use of a full encoder-decoder architecture is not mandatory. The following paragraphs introduce the main models proposed for sequential data.

\paragraph{Recurrent Neural Networks}
Recurrent Neural Networks (RNNs) are employed to detect patterns in sequential data such as text, time series or even genomes \cite{Schmidt2019}. RNNs process an input sequence step-by-step, updating a hidden state at each one. This hidden state retains historical information of the sequence up to the current time step, providing an abstract and summarized representation of the entire input sequence once the last step is processed. It is necessary to address a set of problems concerning these networks when processing long enough series. The loss of long-range dependencies, the vanishing or exploding gradient problem and their sequential nature are the principal ones. Improvements on RNNs have been proposed for mitigating the first two, such as Long Short Term Memory (LSTM) \cite{Hochreiter1997} cells or Gated Recurrent Units (GRU) \cite{ChungGCB14} . 

\paragraph{Convolutional Neural Networks}
Other proposals also aim to address the issue of computing input series sequentially. Convolutional Neural Networks (CNNs) \cite{LeCun1989, wu2017introduction} are characterized by their good performance on image processing, but they have also been used for sequence modeling tasks. The main feature of these networks is the use of convolutions, as their name suggests. Nevertheless, these are not entirely suitable for sequence modeling tasks. These networks may disregard the temporal order of the data, incorporating subsequent observations when processing earlier ones. Temporal Convolutional Networks (TCNs) \cite{Bai2018} combine some of the best practices of modern convolutional architectures for sequence modeling, claiming to outperform RNNs in a broad collection of related tasks. A modification of the baseline convolutional operator originally used for audio generation \cite{denOord2016} is employed, known as causal dilated convolution, inside the main component of these networks. This component is called residual block. The causality of this operator allows the model to respect the temporal order of the data, while the dilatation technique maximizes the coverage of the receptive field for perceiving longer time dependencies without increasing the computational cost or the number of trainable parameters.   

\begin{figure}
    \centering
    \includegraphics[width=0.5\textwidth]{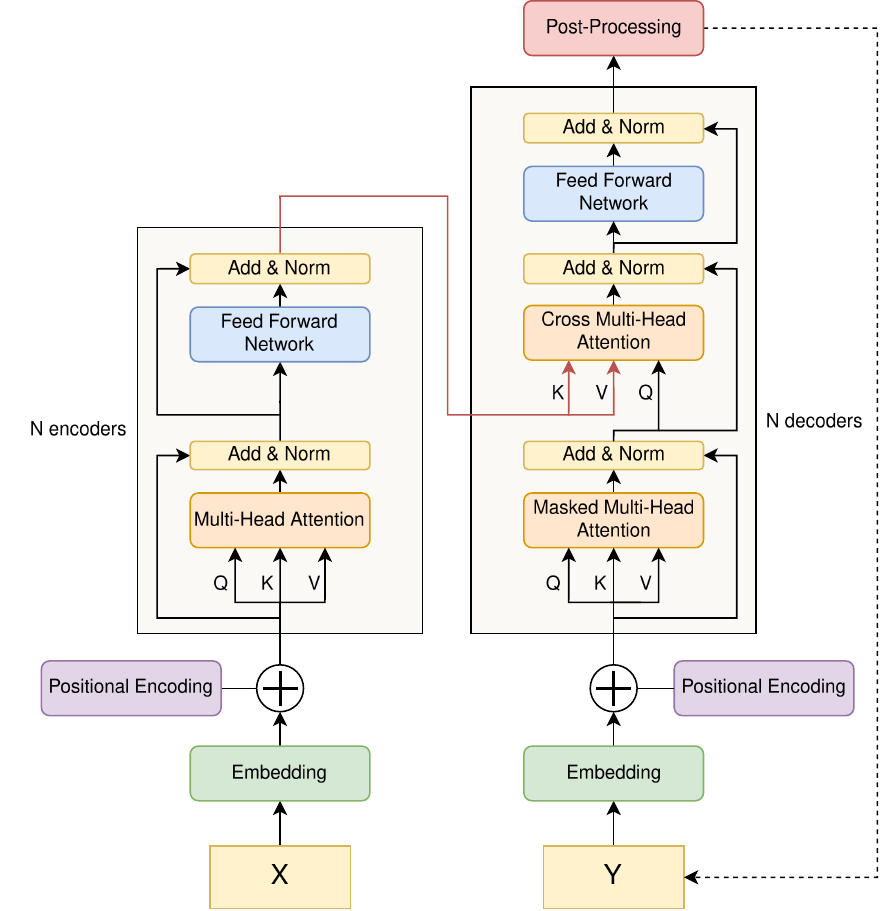}
    \caption{\hl{Detailed baseline Transformer architecture. Encoders capture the relationships between input tokens using multi-head attention and enhance their representations with feed-forward layers. Decoders relate the output of the encoder stack to the predicted tokens using masked and cross-attention mechanisms. Both use positional encoding. Post-processing phase (red block) is usually used to predict the next token according to the task using the output of the decoder stack. Figure adapted from \cite{Vaswani2017attention}}}.
    \label{fig:2.1-transformer-arch}
\end{figure}

\paragraph{Transformers} Transformers \cite{Vaswani2017attention} arose as state-of-the-art architectures for sequence translation with multiple stacked encoders and decoders, as illustrated in Figure \ref{fig:2.1-transformer-arch}, capable of processing the input sequence parallelly and demonstrating superior performance compared to other classic models. Encoders and decoders are composed of Multi-Head Attention (MHA) mechanisms and Feed Forward Networks (FFN) with residual connections and layer normalization. It is important to note that while the encoder stack processes all elements of the input series in a single forward pass, the decoder follows an autoregressive fashion when predicting the output sequence. MHA enhances the capacity of the model by simultaneously computing various attentions, referred to as “heads”, which lets it focus on different aspects of the data in each one. Note that the parallel processing of all tokens entails the need to introduce positional information into their representations to ensure that it takes into account the order of the tokens. The base architecture of Transformers has been adapted for numerous tasks, including TSF. These modifications are highly relevant today due to their strong performance across various applications. The adaptations range from simple single-component changes to entirely new architectures \cite{Cabrera2023}. \hl{Diverse proposals have leveraged Transformer architectures as the primary framework for TSF tasks. Informer \cite{zhou_informer_2021} was one of the first adaptations of this architecture, addressing key issues found in the standard model by introducing a novel attention mechanism that reduces time and memory complexities and using a distillation operation to remove redundant information. DSFormer \cite{yu_dsformer_2023} employs attention mechanisms to fuse information from different levels, effectively capturing significant patterns in both the temporal and feature dimensions of the data, alongside specialized sampling methods. PatchTST \cite{nie_time_2023} handles multivariate time-series by isolating features into separate channels that are divided into patches for either supervised or self-supervised training. These patches are then processed by a Transformer backbone to generate predictions.}

\subsection{Transfer Learning}
\label{subsec:tl}
Using DL methods requires abundant amounts of training and test data from a specific domain to resolve tasks. However, in real-world scenarios, obtaining such data can be challenging or extremely expensive. Therefore, it is worth exploring whether knowledge from a known domain can be applied to an unknown one like humans do. This is the main purpose of Transfer Learning \cite{pan2009survey}.

This approach focuses on improving the predictive function $f_T$ for a specific task $T_T$ within a given domain $D_T$ by using related information from a different task $T_S$ in another domain $D_S$. Essentially, it means borrowing knowledge from a related but different area to enhance the performance of the target task. It is crucial to understand that the source and target domains and tasks may not be the same ($D_S \neq D_T $, $ T_S \neq T_T$), and each one have its own unique set of features and labels \cite{Weiss2016}. 

Originally, models were pre-trained on large datasets for specific time series tasks like classification, forecasting, clustering, anomaly detection, and imputation \cite{Ma2023}. \hl{This setup enabled a relatively easy adaptation of models to new domains within the same task category. Such models are known as Pre-trained Models}. 

FM are an emergent set of models which leverages the capabilities of traditional Pretrained Models, being Transfer Learning one of their fundamental characteristic. Nowadays, a type of FMs called LLMs are widely recognized. These models can abstract various language domains due to the vast amounts of diverse data they are trained on. This generalization ability suggests potential applications beyond language tasks, which will be applied next, such as zero-shot time-series predictors \cite{gruver2023large}. The next paragraphs describe the concept of FM and LLM.

\paragraph{Foundation Model} \hl{The idea of FMs originated with researchers at Stanford University. FMs leverage Transfer Learning and scalability, thanks to advancements in computational power and the wealth of large, diverse datasets. These models act as a robust foundation, adaptable and fine-tunable for an array of tasks, thereby increasing their utility across various fields \cite{bommasani2022opportunities}. A notable characteristic is their exceptional generalization ability, which enables them to perform multiple tasks without additional examples (zero-shot learning) or with minimal examples (few-shot learning), as they can infer from the input context (in-context learning). In essence, FMs signify a paradigm shift in the domain, promoting uniformity with a single model's application to multiple tasks and fostering new functionalities by leveraging their extensive knowledge base.}

\paragraph{Large Language Models} \hl{LLMs represent a specialised category of FMs trained on extensive text data. Their exceptional general-purpose language comprehension equips them to tackle a range of tasks effectively without the need for further fine-tuning}. Currently, most LLM architectures are built on the Transformer model \cite{Vaswani2017attention}. The most prominent models in this domain are developed and supported by large corporations. The LLaMA series, introduced by Meta in 2013, includes several foundational language models that were crafted using an adapted Transformer decoder architecture \cite{touvron2023llama}. These models typically undergo pre-training on a mix of publicly accessible sources in different languages, including Wikipedia, GitHub, and ArXiv. Upgrades to this series have led to the launch of LLaMA-2 later in 2013 \cite{touvron2023llama2} and LLaMA-3 in 2024 \cite{dubey2024llama3herdmodels}, with distinctions mainly in the scale and quality of pretraining data and an expanded context window. Another notable series is the Generative Pre-trained Transformer (GPT) models, created and maintained by OpenAI. Initially formulated as an NLP multitask pre-trained decoder-only transformer \cite{radford_improving_nodate}, these models have evolved into more extensive and sophisticated multimodal systems in their latest iterations, such as GPT-4 \cite{openai2024gpt4technicalreport} and GPT-4o, the latter being capable of handling audio, visual, and textual data in real time. Not all FMs originated as LLMs. For example, Google's Gemini family commenced with a unique decoder-only Transformer multimodal architecture. This structure is adaptable for various applications, from complex reasoning tasks to scenarios with limitations in memory on the device \cite{geminiteam2024geminifamilyhighlycapable}.

\hl{\paragraph{Use of FM for TSF} Due to the similarities between textual and sequential data, various new proposals have arisen to exploit the properties of existing pre-trained and FMs for their adaptation to time-series analysis tasks. Other works opted for the creation of new models using recognized architectures proposing methods to align the representation of time-series with the model original input data. Transformer-based architectures are heavily utilized as the main backbone of these methods \cite{liang2024foundation} as they can handle sequential data effectively, but is not limited to only these. STEP \cite{shao_pre-training_2022} leverages the possibilites of Spatio-Temporal Graph Neural Networks (STGNN) creating a pre-trained time-series model built in a two-stage process. First, a transformer model is pre-trained to generate patches (or segment-level representations) containing relevant contextual information from the input time series using the masked autoenconding strategy. Then, a STGNN leverages these representations to make long-term forecasts of the input series. Other proposals may use LLMs as their base considering the similarities between temporal and textual data. TimeLLM \cite{jin_time-llm_2024} is an LLM reprogramming framework that facilitates the adaptation of these models to the TSF, capable of combining numerical and domain-specific information without requiring their fine-tuning. It introduces two complementary techniques to improve LLMs reasoning capacities related to time-series, Prompt-as-Prefix and Patch Reprogramming. Prompt-as-Prefix enhances the adaptability and guidance of LLMs using a textual description of the dataset domain, statistics and the instruction of the task to be resolved. Patch Reprogramming projects the input window into the LLM data representation space, i.e, match time-series to a combination of several sets of textual tokens presented in the model vocabulary (prototypes) such as "steady down" or "short up". According to the authors, this activates the LLM ability to understand and reason with time series data. Chronos \cite{ansari_chronos_2024} is another framework that employs LLM as its backbone using scaling and quantization techniques to tokenize the time-series, being able to adapt these models without changing anything from their design. This straightforward process achieved decent results, although a full training of the model with large amounts of datasets is needed to achieve them. The Frozen Pre-trained Transformer (FPT) architecture \cite{zhou_one_2023} originated from a unified framework capable of adapting LLMs to diverse Time-Series Analysis tasks. It freezes the MHA and feed-forward layers of the pre-trained model and fine-tune the rest, reducing the cost of training a complete model. However, the output layer needs to be adapted according to the desired task. Lastly, other proposals focus on adapting the model by attempting to guide the LLM correctly on the basis of its input prompt alone, such as Promptcast \cite{Xue2023}. This prompting schema conditions the model to produce a consistent output for the TSF task by introducing information about the data domain, the list of input lags, and a question indicating the number of instants to predict.}

\subsection{Fine-tuning of LLM}
\label{subsec:ft-llm}
\hl{Fine-tuning helps FM achieve optimal performance for specific tasks using small and task-specific data sets. The large number of parameters in these models makes it impractical to modify all of them. Parameter Efficient Fine-Tuning (PEFT) techniques \cite{xu2023parameterefficientfinetuningmethodspretrained, han2024parameterefficientfinetuninglargemodels} allow for efficient and fast fine-tuning of a model by changing only a small and selected number of weights while leaving the rest unchanged.} Although these methods arose mainly for the adaptation of LLMs, they are not limited to them or to the Transformer architecture. These can be grouped into four categories \cite{han2024parameterefficientfinetuninglargemodels}: additive, which modify the model architecture to introduce new trainable components; selective, which refine only a subset of the model weights; re-parameterization, which transform the model in its fitting stage, but then integrate these transformations into its original architecture; and hybrid, which combine several methods. 

\paragraph{Adapters} Adapters \cite{houlsby2019parameterefficienttransferlearningnlp} are a prominent set of methods which introduce small trainable layers between the various modules of the model. Within this category we also find prompt-tuning \cite{lester2021powerscaleparameterefficientprompt}, useful in multitask language models, which seeks to adapt a human input (hard-prompt) to a more affine model (soft-prompt) by operating directly on the continuous vector space given by the model encoder. This overrides the interpretability of the original input. Similar to prompt-tuning are prefix-tuning procedures \cite{li2021prefixtuningoptimizingcontinuousprompts}, which only adds a set of trainable vectors for each task faced by the model at the beginning of the input. 

\paragraph{Low-Rank Adaptations} Low-Rank Adaptations (LoRA) \cite{hu2021loralowrankadaptationlarge}, which fall into the category of re-parameterization. They introduce in parallel in the modules of a model two trainable weight matrices which are combined with each other's output, and which have a rank much smaller than the dimension of the model to be fitted. These matrices can be integrated into the original model parameters (known as the merge stage) ensuring that their inference time does not vary.

\section{LLIAM: Llama Lora-Integrated Autoregressive Model}
\label{sec:lliam}
LLMs are pre-trained with massive amounts of data from different sources. Due to the vast volume of information that a model has to process, significant computational power and time is demanded to build such an extensive knowledge. PEFT techniques enable us to rapidly adapt these models to specific tasks, leveraging their prior knowledge. Our contribution adapts and evaluates the LLM LlaMA for TSF using the LoRA reparametrization technique based on its open-source Lit-LLaMA implementation\footnote{\url{https://github.com/Lightning-AI/lit-llama}} \hl{and a generalisation of the PromptCast \cite{Xue2023} prompting scheme for time-series, which allows us to show how these models, using only the input data of conventional DL algorithms, perform on the TSF task.} The resulting \hl{approach} is called LLIAM: The Llama Lora-Integrated Autoregressive Model.

Our proposal employs the \hl{LLaMA-1} 7B model because it is the most lightweight one. A stacked Transformer decoder-only architecture is employed. Before entering the information into the decoder stack, the input series is transformed \hl{with our general prompting scheme} into a textual prompt and converted into a numerical vector using an embedding. RMS Normalization is applied before processing the data in the Multi-Headed Attention module (MHA) and the Feed Forward Network (FNN) to stabilize the training of the model. MHA allows learning the relationships between the elements of the input data. An efficient implementation of it is used, with a cache that allows the recycling of previous computations to reduce the computational cost. Finally, an FFN is used to improve the features computed by MHA. The output of one decoder is used as the input of the next. Finally, the last output projected with a linear layer, used to predict the next token of the series, which will be added to the input to infer the next one.

\begin{figure}
    \centering
    \includegraphics[width=\textwidth]{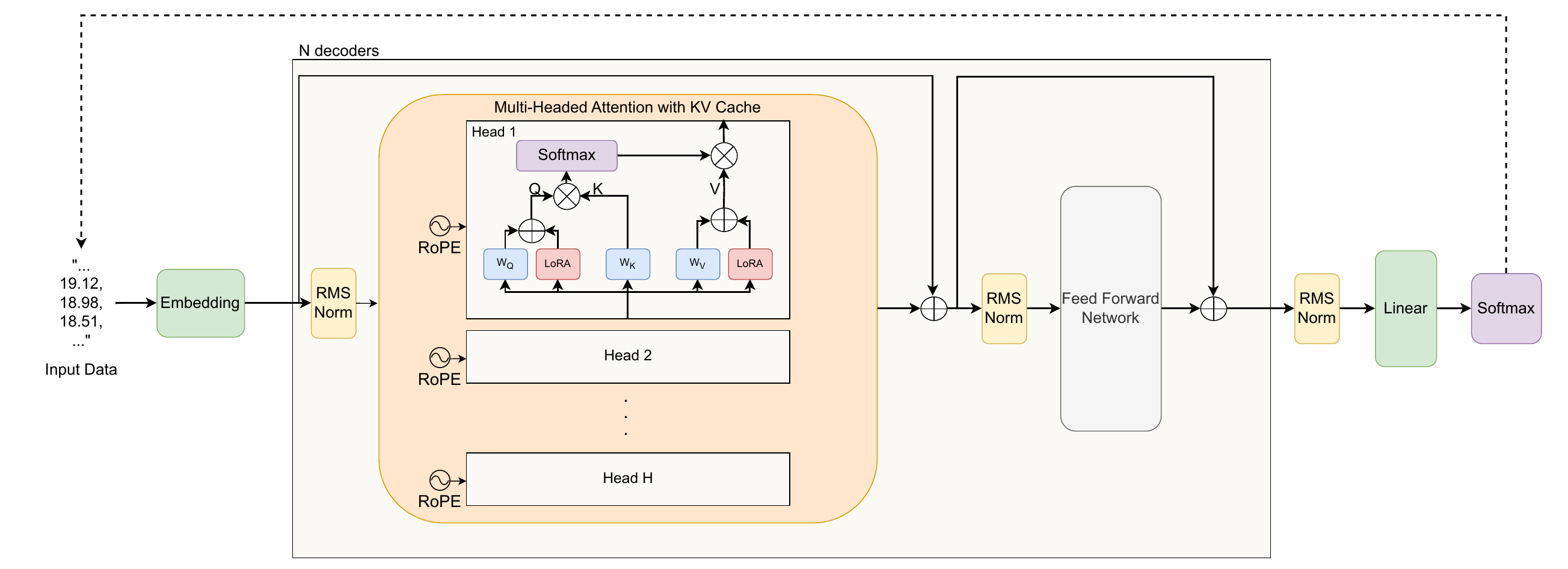}
    \caption{\hl{LLIAM architecture. Only Low-Rank adaptations applied to Query (Q) and Values (V) are trained. The rest of the models is freezed. Our prompting scheme is applied to the numerical input data to align with the textual representation used by the LLM.}}
    \label{fig:lliam-arch}
\end{figure}

In Figure \ref{fig:lliam-arch} the proposed architecture of LLIAM is presented. The following paragraphs describe its main \hl{mechanisms}.

\paragraph{Input Data} \hl{LLaMA-1 7B} can only process textual information as its input as a LLM. Time-series are prompted into the model, following a \hl{simplification of the} prompting scheme Prompcast \cite{Xue2023}. The numerical sequential data is transformed into textual strings following the pattern presented in Table \ref{tab:promptcast-mod}. As seen, our proposal aims for a more generic pattern that does not introduce any exogenous variable or data into the prompt. In contrast with the approach proposed by Promptcast, only the number (\textit{n}) and value of the input observations (\textit{series}) and the desired forecast horizon (\textit{h}) is provided to the model. \hl{It} is expected to only return the desired future values of the series ($t_{n+1}, ..., \hat{t}_{n+h}$). This will enable the use of the same template for all the datasets and evaluate the model output, relying on only its capabilities to extract the different patterns presented in the input series without providing any additional helpful information. \hl{Using a common prompt during the fine-tuning phase of the model helps differentiate between a TSF task and other unrelated tasks. This way, our approach can recognize when it is dealing with a forecasting task.}

\begin{table*}[]
\centering
\begin{tabular}{lll}
\hline
\multirow{3}{*}{Promptcast}   & Context  & From \{$t_1$\} to \{$t_n$\}, the \{variable descriptor\} was \{series\} \{units\} on each \{frequency\} \\  
                              & Question & What is the \{variable descriptor\} going to be on the next \{n+h\} \{frequency\}?                                          \\ 
                              & Answer   & The \{variable descriptor\} will be \{$t_{n+1}, ..., \hat{t}_{n+h}$\} \{units\}.                       \\ \hline\hline
\multirow{3}{*}{LLIAM} & Context  & The last \{n\} observations of an unknown variable were \{series\}.                                   \\ 
                              & Question & \begin{tabular}[c]{@{}l@{}}What will the next \{h\} observations be? \end{tabular} Response:         
                              \\
                              & Answer   & \{$t_{n+1}, ..., \hat{t}_{n+h}$\}                                                     \\ \hline
\end{tabular}%
\caption{Comparison between our prompting scheme and Promptcast, which it is based on. Expected answers are also provided.}
\label{tab:promptcast-mod}
\end{table*}

\paragraph{Embedding} An embedding is used to transform textual tokens into vectors of continuous values in order to ensure a correct computation of the input data by the model. In LLaMA-1 7B, each vector has a dimensionality equal to 4096 \cite{touvron2023llama}. This hyperparameter is known as the dimensionality of the model ($d_m$), and must be consistent across all the modules of LLaMA-1 7B to ensure uniformity between their inputs and outputs. Therefore, all of them will also have $d_m$ dimensions.

\paragraph{RMS Norm} Layer Normalization (LN) \cite{ba2016layernormalization}  is employed to improve and stabilize the training of the baseline Transformer after each of its modules. However, LLaMA-1 7B developers opted for changing what normalization technique is applied and when. Specifically, they applied Root Mean Squared Normalization (RMS Norm) \cite{NEURIPS2019_1e8a1942} before processing the inputs on every module. This modification was inspired by GPT3.

\paragraph{Multi-Headed Attention with KV-Cache} The baseline Scaled Dot-Product Attention \cite{Vaswani2017attention} is used in LLaMA-1 7B. This mechanism is capable of learning relationships between two inputs computationally-efficiently. This procedure can be interpreted as a similarity measure between the different pair of elements that compose the input. In equation \ref{eq:1-attn} is shown how this attention matrix is computed. For a given embedding of the input data $X$, three new representations are calculated, called queries ($Q$), keys ($K$) and values ($V$). Queries represent what an element of $X$ wants to attend, keys show what valuable information a token can provide and $V$ what an element really is. Each of these representations are generated by multiplying three learnable weights matrices to $X$. The similarity between the queries and keys is determined by their dot product, represented as $QK^T$. Higher values indicate greater similarity. A softmax function is applied to produce a probability distribution that describes how much attention each query pays to each key.

\begin{equation}
    Attention(Q, K, V) = softmax(\frac{QK^T}{\sqrt{d_k}})V
    \label{eq:1-attn}
\end{equation}

Nevertheless, employing a single attention mechanism limits the range of relationship a model can notice. Consequently, many of them, known as heads, are computed with different $Q$, $K$ and $V$ representations at the same time. This is referred to as Multi-Headed Attention (MHA) and provides an enhancement of the capacity of the model, enabling the discovery of relationships at different levels. Equation \ref{eq:2-mha} shows how MHA is computed. The final MHA module output is obtained by horizontally concatenating the heads and projecting them into the model’s dimensionality ($d_{model}$) using a linear layer (weights matrix $W_O$). In LLaMA-1 7B, the number of heads used is 32.

\begin{equation}
    \begin{aligned}
        &MHA(X) = Concat(head_1,..., head_h)W^O \\ 
        &\text{where} \quad head_i = Attention(XW_i^Q, XW_i^K, XW_i^V)
        \label{eq:2-mha}
    \end{aligned}
\end{equation}

The autoregressive nature of LLaMA-1 7B leads to repeating redundant calculations that remain unchanged when adding the predicted output tokens to the input. To speed up the computation of MHA, the queries and keys of the input sentence are cached at each step. When predicting the $i^{th}$ token, two auxiliary matrices are used to retrieve these representations from the first to the $(i-1)^{th}$ token. Consequently, only the key and value for the $i^{th}$ element need to be computed instead of the full $Q$ and $V$ matrices. Also, these new values are stored in these caches, facilitating the calculation for the $(i+1)^{th}$ token.

\paragraph{Rotary Positional Embedding (RoPE)} Instead of using the absolute positional encoding proposed by Vaswani et al. \cite{Vaswani2017attention}, LLaMA-1 7B employs a relative positional encoding method known as RoPE \cite{Su2021RoFormer}. This technique integrates positional information into the MHA module using a 2D rotation matrix, applicable to any even-dimensional space. In essence, for an element of the input at position $m$, RoPE rotates the corresponding key and query by $m$ times a fixed angle $\theta$. When the dimensionality of the keys ($d_K$) and queries ($d_V$) are higher than two, dimensions are grouped in $d/2$ pairs and rotated following the set of different rotation angles ($\Theta$) given by equation \ref{eq:rope-angle}. 
\begin{equation}
    \Theta = \{\theta_i = 10000^{-2(i-1)/d}, i \in [1,2, ..., d/2]\}
    \label{eq:rope-angle}
\end{equation}

\paragraph{Feed Forward Network (FFN)} FFNs are used at the end of each decoder to introduce non-linearity while the MHA module processes relationships between positions. The original Transformer architecture uses an FFN with a ReLU activation function. However, LLaMA-1 7B improves performance by using the SwiGLU activation function \cite{Shazeer2020}, which has been shown to outperform Transformers using ReLU on various language understanding tasks.

\paragraph{Low-Rank Adaptations on MHA} As introduced in Section \ref{subsec:ft-llm}, Low-Rank Adaptation (LoRA) \cite{hu2021loralowrankadaptationlarge} is a PEFT technique used for fine-tuning FMs. This method can be incorporated into the model during production deployment, resulting in no additional inference latency, unlike other techniques such as adapters. LoRA was motivated by the conclusions obtained in \cite{aghajanyan2020}, which specify that as models become larger, fewer parameters are needed to fine-tune them to achieve a certain accuracy on a specific task (a concept known as intrinsic dimension). Therefore, the change in the parameters of a model when fine-tuning may also have a low intrinsic dimension. A reparameterization such as LoRA may achieve competitive results in LLMs.

Equation \ref{eq:lora-base} describes how the LoRA technique is added into a DL module. Given a pre-trained model with $W^{model} \in \mathcal{R}^{d_{m} \times k}$ weights, the change needed for fine-tuning it to a certain task is given by $\Delta W$. Based on the previous hypothesis, it must have a low intrinsic dimension. This is ensured by expressing $\Delta W$ as a low-rank decomposition of two matrices $A$ and $B$, where $A \in \mathcal{R}^{r \times k}$ and $B \in \mathcal{R}^{d_{model} \times r}$. $r$ is the hyperparameter responsible for defining the maximum desired rank of these matrices, therefore it must be $r \ll min(k,d)$. It should be noted that $W^{model}$ is not changing during the fine-tuning phase, and only $A$ and $B$ are trainable. The matrix $B$ is initialized to 0 to ensure that $\Delta W=0$ at the beginning of the training.

When processing an input matrix $X$, the output $H$ of a fine-tuned module equals to the combination of its multiplication by the original weights $W^{model}$ and by the specialized adjustment for a task given by $\Delta W$. This last component is scaled by a factor $\alpha/r$, indicating the influence of the LoRA over the output and helps to reduce the need to retune hyperparameters when $r$ is changed. $\alpha$ is set to a fixed value, thus this scale factor is not modified during fine-tuning. 

\begin{equation}
    \begin{aligned}
        &W = W^{model} + \frac{\alpha}{r}\Delta W = W^{model} + \frac{\alpha}{r}(BA)\\ 
        &H = WX = W^{model}X + \frac{\alpha}{r}(BA)X\\ 
        \label{eq:lora-base}
    \end{aligned}
\end{equation}

LoRA can be applied to almost every module of LLaMA-1 7B, \hl{but it is only applied over} the queries and values projections, as illustrated in figure \ref{fig:lliam-arch}.

To conclude this section, Algorithm \ref{alg:forward} presents the pseudo-code for the forward step of LLIAM, offering a clearer understanding of how all the modules interact. For simplicity, the initialization of RoPE and KV-Cache management has been omitted. Within the ATTENTION procedure (line 8), RoPE and LoRAs are applied to Q and V. A forward step of LLIAM produces unnormalized values (logits), which are subsequently normalized using a softmax function in the prediction step outlined in Algorithm \ref{alg:predict}. During this step, LLIAM generates predictions in an autoregressive manner. The stop criterion checks if the maximum length has been reached or if the model has outputted the end of sentence (EOS) token. After generating all the tokens, the textual output is parsed and sanitized to construct the next predicted $h$ values of the input series.
\begin{algorithm}
    \begin{algorithmic}[1]
    \State $n\_heads \gets 32$
    \State $n\_embd \gets 4096$
    \State $n\_layer \gets 32$
    \Procedure {LLIAM-FORWARD} {$X$}
        \State $X_{coded} \gets$ EMBEDDING($X$, $n\_embd$)
        \For {$decoder_i \quad i \in \{1, ..., n\_layer\}$ } %
            \State $X_{norm1} \gets $ RMS-NORM($X_{coded}$)
            \State $X_{attn} \gets $ ATTENTION($X_{norm1}$, $n\_heads$)
            \State $X_{res1} \gets X_{attn} + X_{coded}$
            \State $X_{norm2} \gets $ RMS-NORM($X_{res}$)
            \State $X_{ff} \gets $ FEED-FORWARD($X_{res}$)
            \State $X_{res2} \gets X_{res1} + X_{ff}$
            \State $X_{coded} \gets X_{res2}$
        \EndFor
        
        \State $Y_{norm} \gets$ RMS-NORM($X_{coded}$)
        \State $Y_{logits} \gets$ LINEAR($Y_{norm}$)
        \State \Return $Y_{logits}$
    \EndProcedure
    \caption{LLIAM Forward Pass}\label{alg:forward}
    \end{algorithmic}
\end{algorithm}

\begin{algorithm}
    \begin{algorithmic}[1]
    \State $X \gets "Prompt..."$ \Comment{Textual prompt}
    \Procedure {LLIAM-PREDICT} {$X$}
        \State $Y_{text} \gets X$
        \While {NOT STOP-CRITERION($Y_{text}$)} %
            \State $Y_{logits} \gets$ LLIAM-FORWARD($Y_{text}$)
            \State $Y_{text} \gets Y_{text}$ $\cup$ GET-WORD-SOFTMAX($Y_{logits}$)
        \EndWhile
        \State $Y_{list} \gets$ PARSE-LLIAM-OUTPUT($Y_{text}$)
        \State \Return $Y_{list}$
    \EndProcedure
    \caption{LLIAM Prediction Step}\label{alg:predict}
    \end{algorithmic}
\end{algorithm}

Since LLIAM is based on the LLaMA-1 7B model, most of its hyperparameters are identical to those detailed in \cite{touvron2023llama}. Table \ref{tab:lliam-hyperparams} lists the main hyperparameters of LLIAM. \textit{\#heads}, \textit{\#embd}, and \textit{\#decoders} refer to the base model architecture, indicating the number of heads in the MHA, the embedding dimension, and the number of stacked decoders, respectively. The remaining hyperparameters relate to the LoRA architecture and training. With $r$ set to 8 and $\alpha$ set to 16, the scale factor when merging the model and LoRA outputs is 2. A light dropout \hl{(0.05)} is applied to the LoRA weights to prevent overfitting. The optimizer algorithm used is AdamW \cite{adamw}. A micro-batch training strategy is adopted and a maximum number of training iterations is also defined. To train the LoRAs a supervised procedure is followed. The generated prompt is tokenized and transformed into a nominal representation using the SentencePieces method \cite{kudo2018sentencepiecesimplelanguageindependent}. Then, input and target tokens are separated. The target tokens logits and the model generated logits are used to calculate a cross-entropy loss.  

\begin{table}[]
\centering
\begin{tabular}{ll}
\hline
\textbf{Hyperparameter} & \textbf{Value} \\ \hline
\# heads       & 32     \\
\# embd        & 4096     \\
\# decoders    & 32     \\
$r$             & 8     \\
$\alpha$          & 16    \\
lora\_dropout  & 0.05   \\
learning rate            & $3 \cdot 10^{-4}$  \\
batch size             & 128   \\
micro bs     & 2     \\
optimizer & AdamW \\
max iterations     & 75000 \\ \hline
\end{tabular}
\caption{LLIAM main hyperparameters}
\label{tab:lliam-hyperparams}
\end{table}

\section{Experimentation framework}
\label{sec:exp-framework}

\hl{In this section the design of the experiments done to evaluate our proposal is introduced. The datasets, pre-processing pipeline and metrics employed on the experimentation are described in Subsection \ref{sec:exp-setup}. For clarity, the analysis has been divided in two stages, each designed to demonstrate the capabilities of our LLM-based proposal. The two groups are described below:} 
\hl{\begin{itemize}
    \item Subsection \ref{sec:exp-study} presents a comparative study between our proposal, LLIAM, and a set of traditional DL techniques used in TSF, such as RNNs and TCNs, and TimeLLM, a state-of-the-art LLM-based approach. Non-parametric statistical tests, such as Friedman and Wilcoxon, are also used to determine the existence of significant differences between all these algorithms.
    \item Subsection \ref{sec:zero-shot} shows a zero-shot study between our proposal and a LLM model. It aims to illustrate how the incorporation of LoRAs and the usage of the described prompting scheme enhances the performance of LLMs as time-series forecasters. The datasets used in this experiment are unknown to both models. This is important as we are evaluating the generalization capacity of the two methods, exploiting their ability to face unknown domains and distributions of the data enabled by Transfer Learning.
\end{itemize}}

\subsection{Experimental setup} 
\label{sec:exp-setup}

\hl{For the comparative study} the Monash University time series forecasting repository has been chosen for selecting almost every dataset used in this study \cite{godahewa2021monash}. Of all the available sets, we filtered out all those where the prediction horizon was not clearly specified in the repository and were not uni-variate. Subsequently, we were left with only 7 main sets from this repository (Electricity, M1 Monthly, M1 Quarterly, M3 Monthly, M3 Quarterly, NN5 Daily, NN5 Weekly). \hl{In addition, the Weather and ILI datasets has been included, extracted from \cite{jin_time-llm_2024}. They have been widely used as benchmark for the time-series forecasting task. We provide the description of each of the 9 datasets selected below:}

\begin{itemize}
    \item \textbf{Electricity} (1 Dataset): Contains the electricity consumption of 370 customers in kilowatts from 2011 to 2014. All series have the same length. \hl{The version whose sampling frequency is weekly has been selected.}

    \item \textbf{M1} (2 Datasets): This dataset is composed of a selection of series from one of the first competitions dedicated to its prediction, called M competitions (M, because its creator is called Makridakis). It consists of 617 monthly time series, related to economics, industry, and demography. Not all series have the same number of measurements. \hl{From it, he versions with monthly and quarterly sampling frequency have been selected.}

    \item \textbf{M3} (2 Datasets): Like the M1 dataset, it contains 1428 series from the third edition of the M competitions. They are again not of the same length and add finance-related phenomena to their domains. \hl{From it, the versions that present a monthly and quarterly sampling frequency have been selected.}

    \item \textbf{NN5} (2 Datasets): Dataset from another competition, with 111 series collecting transactions from ATMs located in Great Britain. In this case, the series have the same length. \hl{From it, the versions with a daily and weekly sampling frequency have been selected.}
\hl{
    \item  \textbf{Weather} (1 Dataset): The Weather dataset \cite{jin_time-llm_2024} includes data collected over a year from 21 meteorological stations in Germany, recorded at 10-minute intervals.
}    
\hl{
    \item  \textbf{ILI} (1 Dataset): This dataset contains weekly data on individuals exhibiting symptoms similar to Influenza from 2002 to 2020. The data has been gathered via the U.S. Outpatient Influenza-like Illness Surveillance Network (ILINet) \cite{jin_time-llm_2024}. In this dataset, the primary variable of interest is the total number of patients.
}
\end{itemize}

\hl{For the zero-shot experiment two of the Electricity Transformer Temperature (ETT) datasets \cite{Zhou2020Informer} and the San Francisco Traffic dataset, both extracted from the Monash University time series forecasting repository are considered. The description of these datasets is presented below:}

\begin{itemize}
    \item \textbf{San Francisco Traffic} (1 Dataset): It contains 862 series collecting occupancy rates of freeways located in the San Francisco Bay Area between the years 2015 and 2016. The version whose sampling frequency is weekly has been selected.

    \item \textbf{ETT} (2 Datasets): Three ETT datasets were created using 2 years of data collected from two distinct counties in China. ETTh1 and ETTh2 provide data at a 1-hour granularity. Each data point includes the target value “oil temperature” which is the only feature used in our experimentation, although six additional power load features are also available.
\end{itemize}

\begin{table*}
\centering
\begin{tabular}{@{}lrlrccl@{}}
\toprule
\textbf{Dataset}      & \textbf{\#Series} & \textbf{Frequency} & \textbf{H}             & \textbf{Input size} & \textbf{Same Length?} & \textbf{Used in} \\ \midrule
electricity           & 321               & weekly             & 8                      & 65                  & \cmark & \hl{Comparative study}     \\
m1                    & 617               & monthly            & 18                     & 15                  & \xmark & \hl{Comparative study}     \\
m1                    & 203               & quarterly          & 8                      & 5                   & \xmark & \hl{Comparative study}     \\
m3                    & 1428              & monthly            & 18                     & 15                  & \xmark & \hl{Comparative study}     \\
m3                    & 756               & quarterly          & 8                      & 5                   & \xmark & \hl{Comparative study}     \\
nn5                   & 111               & daily              & 56                     & 9                   & \cmark & \hl{Comparative study}     \\
nn5                   & 111               & weekly             & 8                      & 65                  & \cmark & \hl{Comparative study}     \\
\hl{weather}               & \hl{1}                 & \hl{10 minutes}         & \multicolumn{1}{l}{\hl{96}} & \hl{512}                 & \hl{-}                     & \hl{Comparative study}    \\
\hl{ILI}                   & \hl{1}                 & \hl{weekly}             & \multicolumn{1}{l}{\hl{24}} & \hl{96}                  & \hl{-}                     & \hl{Comparative study}     \\
San Francisco traffic & 862               & weekly             & 8                      & 65                  & \cmark & Zero-shot        \\
ETTh1                 & 1                 & hourly             & 48                     & 24                  & -                     & Zero-shot        \\
ETTh2                 & 1                 & hourly             & 48                     & 24                  & -                     & Zero-shot        \\ \bottomrule
\end{tabular}
\caption{Basic description of the data sets used. The columns \#Series, H, Same Length and Input Size indicate the number of series in the data set, the size of the prediction horizon, the size of the input windows and whether all series have the same length. The study in which each one is used is also indicated.}
\label{tab:desc-datos}
\end{table*}

A sliding window approach is used to adapt the TSF problem as a supervised learning problem. An instance consists of two windows of observations, ordered in time and without gaps. The first represents the input values to be taken by the model ($X$) and the second the expected output ($Y$). Thanks to this restructuring, \hl{series has been transformed} into a set of $(x, y)$ pairs compatible with LLIAM. With this conversion, a textual prompt is constructed following the template described in Section \ref{sec:lliam}, leaving the datasets ready to be used by the model. Table \ref{tab:desc-datos} shows the input window size and the prediction horizon used to construct the sliding windows for each data set, other relevant properties, and in which study each dataset is used. 

\begin{table*}
\centering
\begin{tabular}{lrrrrrr}
\hline
\multicolumn{1}{c}{\textbf{model}} &
  \multicolumn{1}{c}{\textbf{epochs}} &
  \multicolumn{1}{c}{\textbf{layers}} &
  \multicolumn{1}{c}{\textbf{dim\_h}} &
  \multicolumn{1}{c}{\textbf{bs}} &
  \multicolumn{1}{c}{\textbf{lr}} &
  \multicolumn{1}{c}{\textbf{\%dropout}} \\ \hline
RNN-LSTM & 50, 100, 200 & 1, 2, 4 & 32, 64, 128 & 16, 32 & \hl{0.001, 0.01} & 0            \\
RNN-GRU  & 50, 100, 200 & 1, 2, 4 & 32, 64, 128 & 16, 32 & \hl{0.001, 0.01} & 0            \\
TCN      & 50, 100, 200 & 1, 2, 4 & 16, 32, 64  & 16, 32 & \hl{0.001, 0.01} & 0, 0.1, 0.25 \\ \hline
\end{tabular}
\caption{Hyperparameters to be used in the experimentation for each non-foundational model. \textit{dim\_h} refers to the size of the hidden state in the recurrent networks and the number of channels generated by the convolutions in the TCN. \textit{bs} indicates the batch size and \textit{lr} the learning rate of the optimizer.}
\label{tab:hyperparams-models-dl}
\end{table*}

For all the experiments, the inference hyperparameters of LLIAM will be set to default, except the temperature ($T$) with values of 10 and 20. $T$ scales the logits emited by LLIAM, which impacts directly the randomness of the model. Lower temperatures sharpens the distribution described by the output of the model, while higher ones flatten them. 

The evaluation of the models is performed offline. The metrics selected are the Root Mean Square Error \cite{shcherbakov2013survey} (RMSE, equation \ref{eq:rmse}) and the Symmetric Mean Absolute Percentage Error \cite{shcherbakov2013survey} (SMAPE, equation \ref{eq:smape})\hl{. This metric is more suitable for this type of problem because it allows us an easier interpretability of the results by taking into account the magnitude of the data and providing a relative measure unaffected by it. This allows its aggregation and use in statistical tests.}

\begin{equation}
    RMSE = \sqrt{MSE} = \sqrt{\frac{1}{n}\sum_{t=0}^n(y_t - \hat{y_t})^2}
    \label{eq:rmse}
\end{equation}

\hl{
\begin{equation}
    SMAPE = \frac{2}{n}\sum_{t=0}^n\frac{|  y_t - \hat{y_t} |}{| y_t | + | \hat{y_t} |}
    \label{eq:smape}
\end{equation}
}
In the experimental study, \hl{the missing rate (MR) \cite{Xue2023} is also observed}. Given that LLMs are not inherently designed for TSF tasks, \hl{it must be taken into account} for their potential to generate incorrect outputs, known as hallucinations or anomalies. In our work \hl{an output with a horizon shorter than the expected one for each dataset is classified as an anomaly}. If the model outputs a longer horizon, \hl{a trim of the} first $h$ values \hl{is applied} and \hl{the rest are disregarded}. The missing rate is defined in equation \ref{eq:missin-rate} as the difference between the number of set instances $n_{test}$ and the correctly outputted ones $n_{decoded}$ divided by the total size of the test set multiplied by 100. 

\begin{equation}
    MR=\frac{n_{test}-n_{decoded}}{n_{test}} \cdot 100
    \label{eq:missin-rate}
\end{equation}

\subsection{\hl{Comparative study}} 
\label{sec:exp-study}
\hl{In this study our proposal, LLIAM, is compared with various conventional DL algorithms, such as RNNs and TCNs, and TimeLLM, a state-of-the-art reprogramming framework to enable the forecasting of time-series with LLMs presented in Section \ref{subsec:tp}. The present study aims to provide evidence that the general training approach employed in LLIAM surpasses the traditional dataset-specific training paradigm. In contrast to other existing methods, such as TimeLLM, which do not support this level of generalization, LLIAM is designed to capture broader patterns across all datasets. }

\paragraph{\hl{Methodology}}
\label{pg:methodology-comp}
\hl{We have grouped the methodology employed for each type of model evaluated in this study, as not all of them undergo the same training process.} 
\hl{
\begin{itemize}
    \item Traditional DL algorithms (RNNs, TCNs) are trained on a dataset-by-dataset basis with all combinations of hyperparameters listed in Table \ref{tab:hyperparams-models-dl}, being the metrics presented in the tables related to these algorithms averages of all configurations. The purpose of doing this is to be able to obtain a value that corresponds to an average (or "common") case of these algorithms.
    \item TimeLLM is also trained on a dataset-by-dataset basis due to the presence of a final lineal layer that projects the codification emitted by the LLM into the size of the expected horizon \cite{jin_time-llm_2024}. The results presented correspond to the average values of all test instances and not different configurations. We have not performed any hyperparameter optimization because it is our interest to find if a baseline configuration of a LLM-based model can surpass the conventional DL algorithms. We have used the same hyperparameters as in \cite{jin_time-llm_2024} for the Weather and ILI datasets when training TimeLLM. Otherwise, the same hyperparameters as for the M4 competition are used. The number of epochs has been set to 10, given by an heuristic. For our study, the validation set was removed and included in the training set.
    \item LLIAM is trained only one time, combining all train partitions into a whole training dataset. As elements that depend on the length of the series to emit the forecast (i.e. a lineal layer with $H$ units) are not included, a more general model than the DL ones and TimeLLM is builded. TimeLLM has not been trained this way as it is not supported by their architecture, although it is a LLM-based model. Following the premise mentioned earlier, hyperparameter optimization is not performed over LLIAM. We have used a baseline configuration specified in Section \ref{sec:lliam}. 
\end{itemize}
}

\hl{All of the series have been preprocessed for the detection and mitigation of anomalous values using the heuristic described in \cite{Martinez2009}. For the RNNs and TCNs, a Min-Max normalization is applied to incentive the convergence into a local-optima of these methods. For TimeLLM a standardization of the data is done, as seen in its GitHub repository\footnote{https://github.com/KimMeen/Time-LLM}, and a brief description of each dataset has also been incorporated because it is required by its prompting scheme. Both normalization and standarization is reversed when calculating the corresponding metrics. Standardization or normalization techniques on the values of the series are not necessary for LLIAM nor dataset-related information is included.}

\hl{The construction of the training and test partitions is straightforward. For datasets with multiple series the sliding window methodology described in Subsection \ref{sec:exp-setup} is applied. Then, a leave-one-out strategy is employed over Electricity, M1 Monthly, M1 Quarterly, M3 Monthly, M3 Quarterly, NN5 Daily and NN5 Weekly. For the remaining two datasets (Weather and ILI) the number of instant contained in the test partition from \cite{jin_time-llm_2024} is extracted and after that, the sliding window strategy is applied, obtaining the test and train windows. Weather test set has been shortened to the last 10\% windows due to its size.}

\paragraph{\hl{Results}}
\hl{The results obtained confirmed that LLM-based techniques, LLIAM and TimeLLM, outperforms RNNs and TCNs in every dataset. As seen in Tables \ref{tab:rmse-per-ds} and \ref{tab:smape-per-ds}, the difference between these techniques average case and LLM-based ones are significant for both metrics. This suggests that LLMs may be capable of understanding temporal patterns in time-series data. It is important to note that LLIAM does not require the inclusion of any additional domain-specific or dataset-related information, such as univariate statistics, as is necessary for TimeLLM. Furthermore, it does not necessitate the standardization or normalization of the data as is required by all the other models, which serves to reinforce the veracity of this hypothesis.}

\hl{When comparing both LLM-based methods using RSME (Table \ref{tab:rmse-per-ds}) it is easily noticed that both achieve similar performance except on the Electricity dataset. The outstanding results obtained by LLIAM and TimeLLM inform us that the generalization enabled by Transfer Learning and the similitude between textual and time-series data allow LLM models to perform the TSF task, activating the time-series analysis knowledge presented in these type of FMs improved by fine-tuning. It is important to note that RMSE, while commonly used, does not allow for direct comparisons across multiple datasets and is not appropriate for statistical testing unlike SMAPE. Furthermore, it is noteworthy that LLIAM possesses the capacity to execute generic training with all datasets combined, without the necessity for additional information or pre-processing steps, in contrast to the limitations exhibited by TimeLLM.}

\hl{Table \ref{tab:smape-per-ds} presents the results calculated using SMAPE. The results obtained form LLIAM and TimeLLM are comparable across the majority of the data sets. Additionally, in both M1 Monthly and M1 Quarterly, LLIAM scores consistently achive results that are twice as high (in precision) as TimeLLM. In the column 'Avg. of Avgs.' is presented how  that LLIAM with T=10 achieves the best result followed by LLIAM T=20 and then TimeLLM. This achievement reinforces the hypothesis that it is not necessary to implement complex adaptation mechanisms on LLMs to achieve good results.}

\begin{figure}
    \centering
    \includegraphics[width=0.8\linewidth]{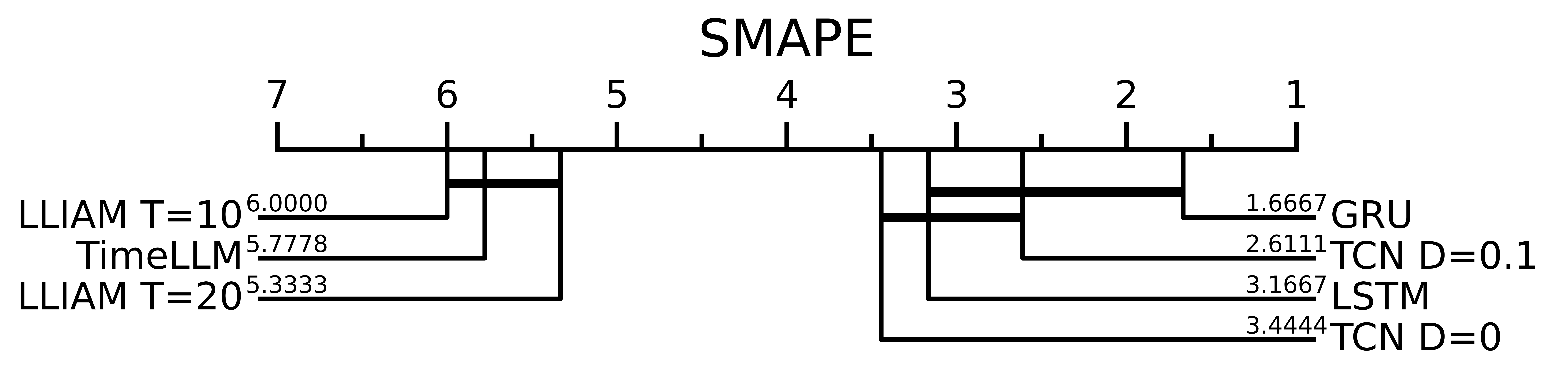}
    \caption{\hl{Critical differences diagram with Wilcoxon post-hoc analysis using the SMAPE metric at $\alpha = 0.05$. Models connected by a horizontal line indicate no statistically significant difference detected by the Wilcoxon test between them. A rank for each model is also indicated next to it.}}
    \label{fig:cd-wilcoxon}
\end{figure}

\hl{To consolidate this analysis, two non-parametric statistical tests using the SMAPE metric have been performed. First, a Friedman test is done, demonstrating that there are significant differences between the results obtained from the different methods evaluated. Then, a paired Wilcoxon test is performed to detect statistical differences between each pair of methods. A critical differences diagram has been made to illustrate better the results obtained after performing these tests. Figure \ref{fig:cd-wilcoxon} illustrates the post-hoc analysis of the Wilcoxon test performed. As expected, no statistical difference has been found between the different LLM-based proposals although a clear gap is set between the conventional DL ones. LLIAM T=10 obtains the highest rank, followed by TimeLLM and LLIAM T=20 which indicates that the adaptation techniques used are capable of achieving competent results.}

\begin{table}[]
\resizebox{\textwidth}{!}{%
\begin{tabular}{@{}lrrrrrrrrr@{}}
\toprule
\multicolumn{10}{c}{RMSE (Lower is better)}                                                                                                                                                                                                                                                                                                                                                                                                                                                                                                                               \\ \midrule
           & \multicolumn{1}{c}{Electricity} & \multicolumn{1}{c}{\begin{tabular}[c]{@{}c@{}}M1\\ Monthly\end{tabular}} & \multicolumn{1}{c}{\begin{tabular}[c]{@{}c@{}}M1\\ Quarterly\end{tabular}} & \multicolumn{1}{c}{\begin{tabular}[c]{@{}c@{}}M3\\ Monthly\end{tabular}} & \multicolumn{1}{c}{\begin{tabular}[c]{@{}c@{}}M3\\ Quarterly\end{tabular}} & \multicolumn{1}{c}{\begin{tabular}[c]{@{}c@{}}NN5\\ Daily\end{tabular}} & \multicolumn{1}{c}{\begin{tabular}[c]{@{}c@{}}NN5\\ Weekly\end{tabular}} & \multicolumn{1}{c}{ILI} & \multicolumn{1}{c}{Weather} \\
LLIAM T=10 & 55293.642                       & 2715.727                                                                 & 2968.888                                                                   & 803.134                                                                  & 634.164                                                                    & 6.279                                                                   & \textbf{18.468}                                                          & 226836.515              & 9.204                       \\
LLIAM T=20 & 53743.873                       & \textbf{2347.011}                                                        & \textbf{2644.296}                                                          & 818.799                                                                  & 639.219                                                                    & 6.296                                                                   & 19.399                                                                   & 217596.959              & \textbf{8.580}              \\
TimeLLM    & \textbf{36781.131}              & 2920.872                                                                 & 2889.753                                                                   & \textbf{801.426}                                                         & \textbf{602.462}                                                           & \textbf{5.677}                                                          & 18.789                                                                   & \textbf{171705.697}     & 10.148                      \\
GRU        & 436904.097                      & 27405.683                                                                & 29961.260                                                                  & 1335.165                                                                 & 1028.065                                                                   & 6.887                                                                   & 24.109                                                                   & 493659.388              & 18.204                      \\
LSTM       & 287614.428                      & 24225.724                                                                & 21713.865                                                                  & 1265.023                                                                 & 964.428                                                                    & 6.677                                                                   & 22.682                                                                   & 492403.464              & 18.860                      \\
TCN D=0    & 670686.818                      & 34761.041                                                                & 13397.682                                                                  & 1326.214                                                                 & 973.804                                                                    & 6.633                                                                   & 22.704                                                                   & 304509.508              & 21.276                      \\
TCN D=0.1  & 663362.613                      & 34764.894                                                                & 15835.914                                                                  & 1347.495                                                                 & 1003.729                                                                   & 6.712                                                                   & 22.721                                                                   & 306990.855              & 21.128                      \\
TCN D=0.25 & 683029.861                      & 32666.035                                                                & 17290.540                                                                  & 1368.059                                                                 & 1001.894                                                                   & 6.863                                                                   & 23.111                                                                   & -                       & -                           \\ \bottomrule
\end{tabular}%
}
\caption{RMSE metric per model and dataset. LSTM, GRU and TCN RMSE are averages for all the configurations tested, meanwhile LLIAM \hl{and TimeLLM} is the average RMSE obtained after aggregating the test instances.}
\label{tab:rmse-per-ds}
\end{table}

\begin{table}[]
\resizebox{\textwidth}{!}{%
\begin{tabular}{@{}lrrrrrrrrrr@{}}
\toprule
\multicolumn{11}{c}{SMAPE (Lower is better)}                                                                                                                                                                                                                                                                                                                                                                                                                                                                                                                                                                                                                      \\ \midrule
           & \multicolumn{1}{c}{Electricity} & \multicolumn{1}{c}{\begin{tabular}[c]{@{}c@{}}M1\\ Monthly\end{tabular}} & \multicolumn{1}{c}{\begin{tabular}[c]{@{}c@{}}M1\\ Quarterly\end{tabular}} & \multicolumn{1}{c}{\begin{tabular}[c]{@{}c@{}}M3\\ Monthly\end{tabular}} & \multicolumn{1}{c}{\begin{tabular}[c]{@{}c@{}}M3\\ Quarterly\end{tabular}} & \multicolumn{1}{c}{\begin{tabular}[c]{@{}c@{}}NN5\\ Daily\end{tabular}} & \multicolumn{1}{c}{\begin{tabular}[c]{@{}c@{}}NN5\\ Weekly\end{tabular}} & \multicolumn{1}{c}{ILI} & \multicolumn{1}{c|}{Weather}         & \multicolumn{1}{c}{\begin{tabular}[c]{@{}c@{}}Avg. \\ of Avgs.\end{tabular}} \\
LLIAM T=10 & 0.0981                          & \textbf{0.1615}                                                          & \textbf{0.1706}                                                            & 0.1472                                                                   & 0.1023                                                                     & 0.2400                                                                  & \textbf{0.1146}                                                          & 0.1661                  & \multicolumn{1}{r|}{\textbf{0.0175}} & \textbf{0.1353}                                                              \\
LLIAM T=20 & \textbf{0.0944}                 & 0.1616                                                                   & 0.1743                                                                     & 0.1495                                                                   & 0.1024                                                                     & 0.2486                                                                  & 0.1178                                                                   & 0.1609                  & \multicolumn{1}{r|}{0.0168}          & 0.1363                                                                       \\
TimeLLM    & 0.1339                          & 0.3483                                                                   & 0.4941                                                                     & \textbf{0.1458}                                                          & \textbf{0.0964}                                                            & \textbf{0.2350}                                                         & 0.1155                                                                   & \textbf{0.1281}         & \multicolumn{1}{r|}{0.0200}          & 0.1908                                                                       \\
GRU        & 0.2410                          & 0.6040                                                                   & 0.6160                                                                     & 0.1650                                                                   & 0.1330                                                                     & 0.2590                                                                  & 0.1350                                                                   & 0.4310                  & \multicolumn{1}{r|}{0.0300}          & 0.2904                                                                       \\
LSTM       & 0.2390                          & 0.6240                                                                   & 0.5790                                                                     & 0.1530                                                                   & 0.1020                                                                     & 0.2470                                                                  & 0.1270                                                                   & 0.4270                  & \multicolumn{1}{r|}{0.0320}          & 0.2811                                                                       \\
TCN D=0    & 0.2420                          & 0.4360                                                                   & 0.4810                                                                     & 0.1600                                                                   & 0.1020                                                                     & 0.2460                                                                  & 0.1290                                                                   & 0.2100                  & \multicolumn{1}{r|}{0.0350}          & 0.2268                                                                       \\
TCN D=0.1  & 0.2320                          & 0.4710                                                                   & 0.4850                                                                     & 0.1660                                                                   & 0.1110                                                                     & 0.2490                                                                  & 0.1290                                                                   & 0.2300                  & \multicolumn{1}{r|}{0.0340}          & 0.2341                                                                       \\
TCN D=0.25 & 0.2470                          & 0.4780                                                                   & 0.5330                                                                     & 0.1680                                                                   & 0.1080                                                                     & 0.2570                                                                  & 0.1320                                                                   & -                       & \multicolumn{1}{r|}{-}               & 0.2747                                                                       \\ \bottomrule
\end{tabular}%
}
\caption{SMAPE metric per model and dataset. LSTM, GRU and TCN SMAPE are averages for all the configurations tested, meanwhile LLIAM \hl{and TimeLLM} is the average SMAPE obtained after aggregating the test instances.}
\label{tab:smape-per-ds}
\end{table}

\begin{table}[]
\resizebox{\textwidth}{!}{%
\begin{tabular}{@{}lrrrrrrrrrr@{}}
\toprule
\multicolumn{11}{c}{Missing Rate \% (Lower is better)}                                                                                                                                                                                                                                                                                                                                                                                                                                                                                                                                                                    \\ \midrule
\multicolumn{1}{c}{} & \multicolumn{1}{c}{Electricity} & \multicolumn{1}{c}{\begin{tabular}[c]{@{}c@{}}M1\\ Monthly\end{tabular}} & \multicolumn{1}{c}{\begin{tabular}[c]{@{}c@{}}M1\\ Quarterly\end{tabular}} & \multicolumn{1}{c}{\begin{tabular}[c]{@{}c@{}}M3\\ Monthly\end{tabular}} & \multicolumn{1}{c}{\begin{tabular}[c]{@{}c@{}}M3\\ Quarterly\end{tabular}} & \multicolumn{1}{c}{\begin{tabular}[c]{@{}c@{}}NN5\\ Daily\end{tabular}} & \multicolumn{1}{c}{\begin{tabular}[c]{@{}c@{}}NN5\\ Weekly\end{tabular}} & \multicolumn{1}{c}{ILI} & \multicolumn{1}{c|}{Weather}        & \multicolumn{1}{c}{Average} \\
LLIAM T=10           & \textbf{0.000}                  & \textbf{0.000}                                                           & \textbf{0.493}                                                             & \textbf{0.000}                                                           & \textbf{0.000}                                                             & 14.414                                                                  & \textbf{0.000}                                                           & \textbf{0.000}          & \multicolumn{1}{r|}{1.313}          & 1.802                       \\
LLIAM T=20           & \textbf{0.000}                  & \textbf{0.000}                                                           & \textbf{0.493}                                                             & \textbf{0.000}                                                           & \textbf{0.000}                                                             & \textbf{7.207}                                                          & \textbf{0.000}                                                           & \textbf{0.000}          & \multicolumn{1}{r|}{\textbf{0.263}} & \textbf{0.885}              \\ \bottomrule
\end{tabular}%
}
\caption{Percentage of test instances unable to decode into the expected forecast horizon. GRU, LSTM, TCN \hl{and TimeLLM} models have been omitted because their missing rate is always 0\%.}
\label{tab:anomalies-percentage}
\end{table}

One \hl{relevant} aspect to \hl{highlight} is the ability of our proposal to produce coherent and expected responses. It must be able to predict at any given time without resorting to making up \hl{anomalies or the so-called LLM hallucinations}. The missing rate for each dataset and LLIAM configuration is reported in Table \ref{tab:anomalies-percentage}. Hallucinations were observed \hl{on three of the nine datasets: NN5 Daily with the highest rate (14.41\% with T=10 and 7.207\% with T=20) followed by Weather (1.313\% with T=10 and 0.263\% with T=20) and lastly M1 Quarterly (0.493\% for both T=10 and T=20)}. The most frequent anomaly is due to the model stopping \hl{generating tokens} early, failing to reach the expected forecast horizon. Early studies showed higher missing rates, but increasing the forecast horizon by one step ($h+1$) reduced this issue. Overall, the total hallucinations are not significant with respect to the results obtained\hl{, achieving an average lower than 2\% with T=10 and lower than 1\% with T=20. As method based on LLM, TimeLLM also generates hallucinations, as is common in such models, but it cannot detect and deal with them properly as they are absorbed by the final projection layer present in its architecture.}

\hl{In regard to the computational cost related to the training of these models, experiments of the methods based on RNNs and TCNs required extensive training times, as all combinations of hyperparameters have to be tried for every dataset. TimeLLM also employs a dataset-by-dataset training procedure, but requires much less training time because it does not require hyperparameter optimization.  LLIAM benefits more from Transfer Learning, enabling the training of a model across all datasets on a single GPU in approximately the same time as TimeLLM. This emphasizes the efficiency of our method in reducing training overhead while maintaining generalization across datasets.}

\subsection{Zero-shot study}
\label{sec:zero-shot}
\hl{In this second study how LLIAM performs on datasets with unknown domains and distributions is tested. We also want to check how the integration of TSF expertise affects the model by comparing the results obtained with respect to its base model, LLaMA-1 7B. TimeLLM could not be included in this study because its architecture has to be trained independently for each dataset, due to an output projection that depends on the number of instants to predict as explained in Section \ref{pg:methodology-comp}.}

\paragraph{\hl{Methodology}} \hl{The weights of LLIAM after training with the comparative study datasets and the LLAMA-1 7B weights provided by Meta are used for inference only. The test partitions of the ETTh1, ETTh2 and San Francisco Traffic datasets are employed, constructed in the same way as explained in the experimental study and applying the anomalies mitigation strategy described above (Section \ref{sec:exp-study}). Due to the large length of the ETTh1 and ETTh2 datasets, the last 10\% of the generated windows to test the models are considered. Each model is evaluated using T=10 and T=20 and the prompting scheme proposed in Section \ref{sec:lliam} is employed in both base model and LLIAM to evaluate only the impact of the new knowledge incorporated over LLIAM with respect LLaMA-1 7B.}

\paragraph{\hl{Results}}\hl{Table \ref{tab:zero-shot-metrics} highlights the average SMAPE and RMSE obtained for each model configuration evaluated. Attention may be drawn by the high average error obtained with LLaMA-1 7B T=20 and T=10 in the dataset ETTh1. This is produced by an anomalous behavior, where the model predicted a very high value in some test windows lags whose tendency is led by low ones, explaining why the SMAPE tends to be much lower. We assume that this happens because LLaMA, although it is capable of emitting forecasts, it does not have the specific knowledge integrated in LLIAM and it may not achieve the intended objective of the task. LLIAM is more robust in comparison with its base model. Generalization on unknown domains is slightly better when using LLIAM instead of LLaMA can also be affirmed, supported by the results obtained. The use of a common prompt template during the training phase of LLIAM achieves its objective of letting the model know when it is facing TSF tasks.}

\begin{table}[]
\centering
\begin{tabular}{@{}lrrr|rrrr@{}}
\toprule
\multicolumn{4}{c|}{RMSE (Lower is better)}                                                                                                               & \multicolumn{4}{c}{SMAPE (Lower is better)}                                                                                                                                \\ \midrule
           & \multicolumn{1}{c}{ETTh1} & \multicolumn{1}{c}{ETTh2} & \multicolumn{1}{c|}{\begin{tabular}[c]{@{}c@{}}San Francisco\\ Traffic\end{tabular}} & \multicolumn{1}{c}{ETTh1} & \multicolumn{1}{c}{ETTh2} & \multicolumn{1}{c|}{\begin{tabular}[c]{@{}c@{}}San Francisco\\ Traffic\end{tabular}} & \multicolumn{1}{c}{Average} \\
LLIAM T=10 & \textbf{2.048}            & 10.176                    & \textbf{1.598}                                                                       & \textbf{0.203}            & 0.236                     & \multicolumn{1}{r|}{\textbf{0.131}}                                                  & \textbf{0.190}              \\
LLIAM T=20 & 2.052                     & \textbf{9.837}            & 1.648                                                                                & 0.216                     & \textbf{0.228}            & \multicolumn{1}{r|}{0.137}                                                           & 0.194                       \\
LLAMA T=10 & 1926800.126               & 701.422                   & 158.113                                                                              & 0.252                     & 0.236                     & \multicolumn{1}{r|}{0.368}                                                           & 0.285                       \\
LLAMA T=20 & 1.93e+111                 & 667.916                   & 159.401                                                                              & 0.247                     & 0.234                     & \multicolumn{1}{r|}{0.372}                                                           & 0.284                       \\ \bottomrule
\end{tabular}
\caption{RMSE and SMAPE metric for each model and dataset on the zero-shot study. It is calculated as the average of all test instances.}
\label{tab:zero-shot-metrics}
\end{table}

\section{Conclusions}
\label{sec:discussion}
\hl{This article introduces a straightforward and general approach for adapting FMs, in particular LLMs, to the TSF task. LLIAM combines the PEFT technique LoRA with a simplification of the PromptCast prompting scheme to improve the performance of LLaMA by integrating task-specific knowledge into the model without increasing its number of parameters. The textual data representation of the input that characterizes LLMs is respected in the application of the prompting scheme. This also serves as a differentiator, helping the LLM to identify when it faces a TSF task. A two-stage experimentation was conducted to demonstrate the capabilities of LLIAM over RNNs, TCNs, another state-of-the-art LLM-base approach, TimeLLM, and the base model employed, LLaMA.}

\hl{In the comparative study it was demonstrated that LLM-based approaches achieve a better performance than the conventional DL ones without the need for intense hyperparameter-tuning. This find suggests that there is a certain similarity between language and time-series that LLM-based models are able to exploit, taking advantage of the knowledge transfer that comes from being pre-trained with a large and heterogeneous amount of data. The straightforward modifications applied to LLIAM are sufficient for enhancing the capabilities of LLMs, as it achieves similar behaviour to TimeLLM and outperforms it with a fairer metric such as SMAPE. Furthermore, the dataset-specific training of TimeLLM, its reliance on contextual information, and its pre-processing are a disadvantage with respect to the one-pass approach of LLIAM. This demonstration serves to illustrate that the methodology employed in this study produces a more general model. The results obtained in the zero-shot study confirm to us that the addition of specialized knowledge related to TSF problems enhances the performance of LLaMA-1 7B, highlighting its generalization capacity over unknown datasets and the robustness of the predictions emitted. TimeLLM is not compatible with the zero-shot study design used in this paper because it needs to be trained with information from the dataset to be predicted in addition to the limitations of its architecture.}

\hl{The findings of the present study have provided answers to the questions that were raised at the initiation of the research. It has been demonstrated that LLMs are capable of leveraging the similarity of text and time-series by implementing easy modifications to them, achieving more general and straightforward solutions than other methods.}

\hl{This work sets the stage for future research involving more extensive testing with different LLMs, including adaptations or architectures specifically tailored for TSF tasks and using larger models, but respecting the original architecture of these models and trying to modify it as little as possible. In addition, the exploration and development of PEFT techniques beyond LoRAs is a promising direction for improving the efficiency and adaptability of these models. Another field to explore is how time-series data have to be inputted into this type of models, including determining what dataset and domain specific information is needed to activate the capabilities of these models as time-series forecasters.} Overall, the arrival of the FMs opens up the possibility of numerous interesting proposals.

\section*{Acknowledgments}

The research carried out in this study is part of the project ``Advances in the development of trustworthy AI models to contribute to the adoption and use of responsible AI in healthcare (TAIH)'' with code PID2023-149511OB-I00 \hl{and supported via the FPU (Formación de Profesorado Universitario) fellowship program, both funded by the Spanish Ministry of Science, Innovation and Universities.}

\bibliographystyle{elsarticle-num-names.bst}

\begin{thebibliography}{56}
\expandafter\ifx\csname natexlab\endcsname\relax\def\natexlab#1{#1}\fi
\providecommand{\url}[1]{\texttt{#1}}
\providecommand{\href}[2]{#2}
\providecommand{\path}[1]{#1}
\providecommand{\DOIprefix}{doi:}
\providecommand{\ArXivprefix}{arXiv:}
\providecommand{\URLprefix}{URL: }
\providecommand{\Pubmedprefix}{pmid:}
\providecommand{\doi}[1]{\href{http://dx.doi.org/#1}{\path{#1}}}
\providecommand{\Pubmed}[1]{\href{pmid:#1}{\path{#1}}}
\providecommand{\bibinfo}[2]{#2}
\ifx\xfnm\relax \def\xfnm[#1]{\unskip,\space#1}\fi
\bibitem[{Zhou et~al.(2023)Zhou, Li, Li, Yu, Liu, Wang, Zhang, Ji, Yan, He, Peng, Li, Wu, Liu, Xie, Xiong, Pei, Yu, and Sun}]{zhou2023comprehensivesurveypretrainedfoundation}
\bibinfo{author}{C.~Zhou}, \bibinfo{author}{Q.~Li}, \bibinfo{author}{C.~Li}, \bibinfo{author}{J.~Yu}, \bibinfo{author}{Y.~Liu}, \bibinfo{author}{G.~Wang}, \bibinfo{author}{K.~Zhang}, \bibinfo{author}{C.~Ji}, \bibinfo{author}{Q.~Yan}, \bibinfo{author}{L.~He}, \bibinfo{author}{H.~Peng}, \bibinfo{author}{J.~Li}, \bibinfo{author}{J.~Wu}, \bibinfo{author}{Z.~Liu}, \bibinfo{author}{P.~Xie}, \bibinfo{author}{C.~Xiong}, \bibinfo{author}{J.~Pei}, \bibinfo{author}{P.~S. Yu}, \bibinfo{author}{L.~Sun}, \bibinfo{title}{A comprehensive survey on pretrained foundation models: A history from bert to chatgpt}, \bibinfo{year}{2023}. \URLprefix \url{https://arxiv.org/abs/2302.09419}. \href{http://arxiv.org/abs/2302.09419}{{\tt arXiv:2302.09419}}.
\bibitem[{Feuerriegel et~al.(2024)Feuerriegel, Hartmann, Janiesch, and Zschech}]{feuerriegel2024generative}
\bibinfo{author}{S.~Feuerriegel}, \bibinfo{author}{J.~Hartmann}, \bibinfo{author}{C.~Janiesch}, \bibinfo{author}{P.~Zschech},
\newblock \bibinfo{title}{Generative ai},
\newblock \bibinfo{journal}{Business \& Information Systems Engineering} \bibinfo{volume}{66} (\bibinfo{year}{2024}) \bibinfo{pages}{111--126}.
\bibitem[{Weiss et~al.(2016)Weiss, Khoshgoftaar, and Wang}]{Weiss2016}
\bibinfo{author}{K.~Weiss}, \bibinfo{author}{T.~M. Khoshgoftaar}, \bibinfo{author}{D.~Wang},
\newblock \bibinfo{title}{A survey of transfer learning},
\newblock \bibinfo{journal}{Journal of Big Data} \bibinfo{volume}{3} (\bibinfo{year}{2016}) \bibinfo{pages}{9}. \URLprefix \url{https://doi.org/10.1186/s40537-016-0043-6}. \DOIprefix\doi{10.1186/s40537-016-0043-6}.
\bibitem[{Jin et~al.(2024)Jin, Zhang, Chen, Zhang, Liang, Yang, Wang, Pan, and Wen}]{jin2024positionlargelanguagemodels}
\bibinfo{author}{M.~Jin}, \bibinfo{author}{Y.~Zhang}, \bibinfo{author}{W.~Chen}, \bibinfo{author}{K.~Zhang}, \bibinfo{author}{Y.~Liang}, \bibinfo{author}{B.~Yang}, \bibinfo{author}{J.~Wang}, \bibinfo{author}{S.~Pan}, \bibinfo{author}{Q.~Wen}, \bibinfo{title}{Position: What can large language models tell us about time series analysis}, \bibinfo{year}{2024}. \URLprefix \url{https://arxiv.org/abs/2402.02713}. \href{http://arxiv.org/abs/2402.02713}{{\tt arXiv:2402.02713}}.
\bibitem[{Gruver et~al.(2023)Gruver, Finzi, Qiu, and Wilson}]{gruver2023large}
\bibinfo{author}{N.~Gruver}, \bibinfo{author}{M.~Finzi}, \bibinfo{author}{S.~Qiu}, \bibinfo{author}{A.~G. Wilson}, \bibinfo{title}{Large language models are zero-shot time series forecasters}, \bibinfo{year}{2023}. \href{http://arxiv.org/abs/2310.07820}{{\tt arXiv:2310.07820}}.
\bibitem[{Touvron et~al.(2023)Touvron, Lavril, Izacard, Martinet, Lachaux, Lacroix, Rozière, Goyal, Hambro, Azhar, Rodriguez, Joulin, Grave, and Lample}]{touvron2023llama}
\bibinfo{author}{H.~Touvron}, \bibinfo{author}{T.~Lavril}, \bibinfo{author}{G.~Izacard}, \bibinfo{author}{X.~Martinet}, \bibinfo{author}{M.-A. Lachaux}, \bibinfo{author}{T.~Lacroix}, \bibinfo{author}{B.~Rozière}, \bibinfo{author}{N.~Goyal}, \bibinfo{author}{E.~Hambro}, \bibinfo{author}{F.~Azhar}, \bibinfo{author}{A.~Rodriguez}, \bibinfo{author}{A.~Joulin}, \bibinfo{author}{E.~Grave}, \bibinfo{author}{G.~Lample}, \bibinfo{title}{Llama: Open and efficient foundation language models}, \bibinfo{year}{2023}. \href{http://arxiv.org/abs/2302.13971}{{\tt arXiv:2302.13971}}.
\bibitem[{Hu et~al.(2021)Hu, Shen, Wallis, Allen-Zhu, Li, Wang, Wang, and Chen}]{hu2021loralowrankadaptationlarge}
\bibinfo{author}{E.~J. Hu}, \bibinfo{author}{Y.~Shen}, \bibinfo{author}{P.~Wallis}, \bibinfo{author}{Z.~Allen-Zhu}, \bibinfo{author}{Y.~Li}, \bibinfo{author}{S.~Wang}, \bibinfo{author}{L.~Wang}, \bibinfo{author}{W.~Chen}, \bibinfo{title}{Lora: Low-rank adaptation of large language models}, \bibinfo{year}{2021}. \URLprefix \url{https://arxiv.org/abs/2106.09685}. \href{http://arxiv.org/abs/2106.09685}{{\tt arXiv:2106.09685}}.
\bibitem[{Bolón-Canedo et~al.(2024)Bolón-Canedo, Morán-Fernández, Cancela, and Alonso-Betanzos}]{BOLONCANEDO2024128096}
\bibinfo{author}{V.~Bolón-Canedo}, \bibinfo{author}{L.~Morán-Fernández}, \bibinfo{author}{B.~Cancela}, \bibinfo{author}{A.~Alonso-Betanzos},
\newblock \bibinfo{title}{A review of green artificial intelligence: Towards a more sustainable future},
\newblock \bibinfo{journal}{Neurocomputing} \bibinfo{volume}{599} (\bibinfo{year}{2024}) \bibinfo{pages}{128096}. \URLprefix \url{https://www.sciencedirect.com/science/article/pii/S0925231224008671}. \DOIprefix\doi{https://doi.org/10.1016/j.neucom.2024.128096}.
\bibitem[{Brockwell and Davis(2016{\natexlab{a}})}]{brockwell2016c8}
\bibinfo{author}{P.~J. Brockwell}, \bibinfo{author}{R.~A. Davis}, \bibinfo{title}{Introduction to Time Series and Forecasting}, Springer Texts in Statistics, \bibinfo{publisher}{Springer International Publishing}, \bibinfo{year}{2016}{\natexlab{a}}. \URLprefix \url{https://doi.org/10.1007/978-3-319-29854-2}. \DOIprefix\doi{10.1007/978-3-319-29854-2}.
\bibitem[{Brockwell and Davis(2016{\natexlab{b}})}]{brockwell2016c1}
\bibinfo{author}{P.~J. Brockwell}, \bibinfo{author}{R.~A. Davis}, \bibinfo{title}{Introduction to Time Series and Forecasting}, Springer Texts in Statistics, \bibinfo{publisher}{Springer International Publishing}, \bibinfo{year}{2016}{\natexlab{b}}. \URLprefix \url{https://doi.org/10.1007/978-3-319-29854-2}. \DOIprefix\doi{10.1007/978-3-319-29854-2}.
\bibitem[{Nielsen(2015)}]{Nielsen2018C4}
\bibinfo{author}{M.~A. Nielsen}, \bibinfo{title}{Neural networks and deep learning}, \bibinfo{year}{2015}. \URLprefix \url{http://neuralnetworksanddeeplearning.com/}.
\bibitem[{Prince(2023)}]{prince2023understanding}
\bibinfo{author}{S.~J. Prince}, \bibinfo{title}{Understanding Deep Learning}, \bibinfo{publisher}{The MIT Press}, \bibinfo{year}{2023}. \URLprefix \url{http://udlbook.com}.
\bibitem[{Sarker(2021)}]{Sarker2021}
\bibinfo{author}{I.~H. Sarker},
\newblock \bibinfo{title}{Deep learning: A comprehensive overview on techniques, taxonomy, applications and research directions},
\newblock \bibinfo{journal}{SN Computer Science} \bibinfo{volume}{2} (\bibinfo{year}{2021}) \bibinfo{pages}{420}. \URLprefix \url{https://doi.org/10.1007/s42979-021-00815-1}. \DOIprefix\doi{10.1007/s42979-021-00815-1}.
\bibitem[{Neubig(2017)}]{Neubig17}
\bibinfo{author}{G.~Neubig},
\newblock \bibinfo{title}{Neural machine translation and sequence-to-sequence models: {A} tutorial},
\newblock \bibinfo{journal}{CoRR} \bibinfo{volume}{abs/1703.01619} (\bibinfo{year}{2017}). \URLprefix \url{http://arxiv.org/abs/1703.01619}. \href{http://arxiv.org/abs/1703.01619}{{\tt arXiv:1703.01619}}.
\bibitem[{Vaswani et~al.(2017)}]{Vaswani2017attention}
\bibinfo{author}{A.~Vaswani}, et~al.,
\newblock \bibinfo{title}{Attention is all you need},
\newblock \bibinfo{journal}{Proc. of NIPS} \bibinfo{volume}{30} (\bibinfo{year}{2017}) \bibinfo{pages}{5999--6009}.
\bibitem[{Schmidt(2019)}]{Schmidt2019}
\bibinfo{author}{R.~M. Schmidt}, \bibinfo{title}{Recurrent {Neural} {Networks} ({RNNs}): {A} gentle {Introduction} and {Overview}}, \bibinfo{year}{2019}. \URLprefix \url{http://arxiv.org/abs/1912.05911}. \DOIprefix\doi{10.48550/arXiv.1912.05911}, \bibinfo{note}{arXiv:1912.05911 [cs, stat]}.
\bibitem[{Hochreiter and Schmidhuber(1997)}]{Hochreiter1997}
\bibinfo{author}{S.~Hochreiter}, \bibinfo{author}{J.~Schmidhuber},
\newblock \bibinfo{title}{Long short-term memory},
\newblock \bibinfo{journal}{Neural computation} \bibinfo{volume}{9} (\bibinfo{year}{1997}) \bibinfo{pages}{1735--80}. \DOIprefix\doi{10.1162/neco.1997.9.8.1735}.
\bibitem[{Chung et~al.(2014)Chung, G{\"{u}}l{\c{c}}ehre, Cho, and Bengio}]{ChungGCB14}
\bibinfo{author}{J.~Chung}, \bibinfo{author}{{\c{C}}.~G{\"{u}}l{\c{c}}ehre}, \bibinfo{author}{K.~Cho}, \bibinfo{author}{Y.~Bengio},
\newblock \bibinfo{title}{Empirical evaluation of gated recurrent neural networks on sequence modeling},
\newblock \bibinfo{journal}{CoRR} \bibinfo{volume}{abs/1412.3555} (\bibinfo{year}{2014}). \URLprefix \url{http://arxiv.org/abs/1412.3555}. \href{http://arxiv.org/abs/1412.3555}{{\tt arXiv:1412.3555}}.
\bibitem[{LeCun et~al.(1989)LeCun, Boser, Denker, Henderson, Howard, Hubbard, and Jackel}]{LeCun1989}
\bibinfo{author}{Y.~LeCun}, \bibinfo{author}{B.~Boser}, \bibinfo{author}{J.~S. Denker}, \bibinfo{author}{D.~Henderson}, \bibinfo{author}{R.~E. Howard}, \bibinfo{author}{W.~Hubbard}, \bibinfo{author}{L.~D. Jackel},
\newblock \bibinfo{title}{{Backpropagation Applied to Handwritten Zip Code Recognition}},
\newblock \bibinfo{journal}{Neural Computation} \bibinfo{volume}{1} (\bibinfo{year}{1989}) \bibinfo{pages}{541--551}. \URLprefix \url{https://doi.org/10.1162/neco.1989.1.4.541}. \DOIprefix\doi{10.1162/neco.1989.1.4.541}.
\bibitem[{Wu(2017)}]{wu2017introduction}
\bibinfo{author}{J.~Wu},
\newblock \bibinfo{title}{Introduction to convolutional neural networks},
\newblock \bibinfo{journal}{National Key Lab for Novel Software Technology. Nanjing University. China} \bibinfo{volume}{5} (\bibinfo{year}{2017}) \bibinfo{pages}{495}.
\bibitem[{Bai et~al.(2018)Bai, Kolter, and Koltun}]{Bai2018}
\bibinfo{author}{S.~Bai}, \bibinfo{author}{J.~Z. Kolter}, \bibinfo{author}{V.~Koltun},
\newblock \bibinfo{title}{An empirical evaluation of generic convolutional and recurrent networks for sequence modeling},
\newblock \bibinfo{journal}{CoRR} \bibinfo{volume}{abs/1803.01271} (\bibinfo{year}{2018}). \URLprefix \url{http://arxiv.org/abs/1803.01271}. \DOIprefix\doi{doi.org/10.48550/arXiv.1803.01271}. \href{http://arxiv.org/abs/1803.01271}{{\tt arXiv:1803.01271}}.
\bibitem[{van~den Oord et~al.(2016)van~den Oord, Dieleman, Zen, Simonyan, Vinyals, Graves, Kalchbrenner, Senior, and Kavukcuoglu}]{denOord2016}
\bibinfo{author}{A.~van~den Oord}, \bibinfo{author}{S.~Dieleman}, \bibinfo{author}{H.~Zen}, \bibinfo{author}{K.~Simonyan}, \bibinfo{author}{O.~Vinyals}, \bibinfo{author}{A.~Graves}, \bibinfo{author}{N.~Kalchbrenner}, \bibinfo{author}{A.~W. Senior}, \bibinfo{author}{K.~Kavukcuoglu},
\newblock \bibinfo{title}{Wavenet: {A} generative model for raw audio},
\newblock \bibinfo{journal}{CoRR} \bibinfo{volume}{abs/1609.03499} (\bibinfo{year}{2016}). \URLprefix \url{http://arxiv.org/abs/1609.03499}. \href{http://arxiv.org/abs/1609.03499}{{\tt arXiv:1609.03499}}.
\bibitem[{Cabrera-Bermejo et~al.(2023)Cabrera-Bermejo, Del~Jesus, Rivera, Elizondo, Charte, and P{\'e}rez-Godoy}]{Cabrera2023}
\bibinfo{author}{M.~I. Cabrera-Bermejo}, \bibinfo{author}{M.~J. Del~Jesus}, \bibinfo{author}{A.~J. Rivera}, \bibinfo{author}{D.~Elizondo}, \bibinfo{author}{F.~Charte}, \bibinfo{author}{M.~D. P{\'e}rez-Godoy},
\newblock \bibinfo{title}{Analysis of transformer model applications},
\newblock in: \bibinfo{booktitle}{Hybrid Artificial Intelligent Systems}, \bibinfo{publisher}{Springer Nature Switzerland}, \bibinfo{address}{Cham}, \bibinfo{year}{2023}, pp. \bibinfo{pages}{231--243}. \DOIprefix\doi{10.1007/978-3-031-40725-3_20}.
\bibitem[{Zhou et~al.(2021)Zhou, Zhang, Peng, Zhang, Li, Xiong, and Zhang}]{zhou_informer_2021}
\bibinfo{author}{H.~Zhou}, \bibinfo{author}{S.~Zhang}, \bibinfo{author}{J.~Peng}, \bibinfo{author}{S.~Zhang}, \bibinfo{author}{J.~Li}, \bibinfo{author}{H.~Xiong}, \bibinfo{author}{W.~Zhang}, \bibinfo{title}{Informer: {Beyond} {Efficient} {Transformer} for {Long} {Sequence} {Time}-{Series} {Forecasting}}, \bibinfo{year}{2021}. \URLprefix \url{http://arxiv.org/abs/2012.07436}. \DOIprefix\doi{10.48550/arXiv.2012.07436}, \bibinfo{note}{arXiv:2012.07436 [cs]}.
\bibitem[{Yu et~al.(2023)Yu, Wang, Shao, Sun, Wu, and Xu}]{yu_dsformer_2023}
\bibinfo{author}{C.~Yu}, \bibinfo{author}{F.~Wang}, \bibinfo{author}{Z.~Shao}, \bibinfo{author}{T.~Sun}, \bibinfo{author}{L.~Wu}, \bibinfo{author}{Y.~Xu}, \bibinfo{title}{{DSformer}: {A} {Double} {Sampling} {Transformer} for {Multivariate} {Time} {Series} {Long}-term {Prediction}}, \bibinfo{year}{2023}. \URLprefix \url{http://arxiv.org/abs/2308.03274}. \DOIprefix\doi{10.48550/arXiv.2308.03274}, \bibinfo{note}{arXiv:2308.03274 [cs]}.
\bibitem[{Nie et~al.(2023)Nie, Nguyen, Sinthong, and Kalagnanam}]{nie_time_2023}
\bibinfo{author}{Y.~Nie}, \bibinfo{author}{N.~H. Nguyen}, \bibinfo{author}{P.~Sinthong}, \bibinfo{author}{J.~Kalagnanam}, \bibinfo{title}{A {Time} {Series} is {Worth} 64 {Words}: {Long}-term {Forecasting} with {Transformers}}, \bibinfo{year}{2023}. \URLprefix \url{http://arxiv.org/abs/2211.14730}. \DOIprefix\doi{10.48550/arXiv.2211.14730}, \bibinfo{note}{arXiv:2211.14730 [cs]}.
\bibitem[{Pan and Yang(2009)}]{pan2009survey}
\bibinfo{author}{S.~J. Pan}, \bibinfo{author}{Q.~Yang},
\newblock \bibinfo{title}{A survey on transfer learning},
\newblock \bibinfo{journal}{IEEE Transactions on knowledge and data engineering} \bibinfo{volume}{22} (\bibinfo{year}{2009}) \bibinfo{pages}{1345--1359}.
\bibitem[{Ma et~al.(2023)Ma, Liu, Zheng, Huang, Zhu, Yu, and Kwok}]{Ma2023}
\bibinfo{author}{Q.~Ma}, \bibinfo{author}{Z.~Liu}, \bibinfo{author}{Z.~Zheng}, \bibinfo{author}{Z.~Huang}, \bibinfo{author}{S.~Zhu}, \bibinfo{author}{Z.~Yu}, \bibinfo{author}{J.~T. Kwok}, \bibinfo{title}{A survey on time-series pre-trained models}, \bibinfo{year}{2023}. \href{http://arxiv.org/abs/2305.10716}{{\tt arXiv:2305.10716}}.
\bibitem[{et~al.(2022)}]{bommasani2022opportunities}
\bibinfo{author}{R.~B. et~al.}, \bibinfo{title}{On the opportunities and risks of foundation models}, \bibinfo{year}{2022}. \href{http://arxiv.org/abs/2108.07258}{{\tt arXiv:2108.07258}}.
\bibitem[{Touvron et~al.(2023)Touvron, Martin, and et~al.}]{touvron2023llama2}
\bibinfo{author}{H.~Touvron}, \bibinfo{author}{L.~Martin}, \bibinfo{author}{K.~S. et~al.}, \bibinfo{title}{Llama 2: Open foundation and fine-tuned chat models}, \bibinfo{year}{2023}. \href{http://arxiv.org/abs/2307.09288}{{\tt arXiv:2307.09288}}.
\bibitem[{Dubey et~al.(2024)Dubey, Jauhri, and et~al.}]{dubey2024llama3herdmodels}
\bibinfo{author}{A.~Dubey}, \bibinfo{author}{A.~Jauhri}, \bibinfo{author}{et~al.}, \bibinfo{title}{The llama 3 herd of models}, \bibinfo{year}{2024}. \URLprefix \url{https://arxiv.org/abs/2407.21783}. \href{http://arxiv.org/abs/2407.21783}{{\tt arXiv:2407.21783}}.
\bibitem[{Radford et~al.(2018)Radford, Narasimhan, Salimans, and Sutskever}]{radford_improving_nodate}
\bibinfo{author}{A.~Radford}, \bibinfo{author}{K.~Narasimhan}, \bibinfo{author}{T.~Salimans}, \bibinfo{author}{I.~Sutskever},
\newblock \bibinfo{title}{Improving {Language} {Understanding} by {Generative} {Pre}-{Training}}  (\bibinfo{year}{2018}).
\bibitem[{OpenAI et~al.(2024)OpenAI, Achiam, Adler, and et~al.}]{openai2024gpt4technicalreport}
\bibinfo{author}{OpenAI}, \bibinfo{author}{J.~Achiam}, \bibinfo{author}{S.~Adler}, \bibinfo{author}{S.~A. et~al.}, \bibinfo{title}{Gpt-4 technical report}, \bibinfo{year}{2024}. \URLprefix \url{https://arxiv.org/abs/2303.08774}. \href{http://arxiv.org/abs/2303.08774}{{\tt arXiv:2303.08774}}.
\bibitem[{Team et~al.(2024)Team, Anil, Borgeaud, and et~al.}]{geminiteam2024geminifamilyhighlycapable}
\bibinfo{author}{G.~Team}, \bibinfo{author}{R.~Anil}, \bibinfo{author}{S.~Borgeaud}, \bibinfo{author}{J.-B.~A. et~al.}, \bibinfo{title}{Gemini: A family of highly capable multimodal models}, \bibinfo{year}{2024}. \URLprefix \url{https://arxiv.org/abs/2312.11805}. \href{http://arxiv.org/abs/2312.11805}{{\tt arXiv:2312.11805}}.
\bibitem[{Liang et~al.(2024)Liang, Wen, Nie, Jiang, Jin, Song, Pan, and Wen}]{liang2024foundation}
\bibinfo{author}{Y.~Liang}, \bibinfo{author}{H.~Wen}, \bibinfo{author}{Y.~Nie}, \bibinfo{author}{Y.~Jiang}, \bibinfo{author}{M.~Jin}, \bibinfo{author}{D.~Song}, \bibinfo{author}{S.~Pan}, \bibinfo{author}{Q.~Wen},
\newblock \bibinfo{title}{Foundation models for time series analysis: A tutorial and survey},
\newblock in: \bibinfo{booktitle}{Proceedings of the 30th ACM SIGKDD Conference on Knowledge Discovery and Data Mining}, \bibinfo{year}{2024}, pp. \bibinfo{pages}{6555--6565}.
\bibitem[{Shao et~al.(2022)Shao, Zhang, Wang, and Xu}]{shao_pre-training_2022}
\bibinfo{author}{Z.~Shao}, \bibinfo{author}{Z.~Zhang}, \bibinfo{author}{F.~Wang}, \bibinfo{author}{Y.~Xu},
\newblock \bibinfo{title}{Pre-training {Enhanced} {Spatial}-temporal {Graph} {Neural} {Network} for {Multivariate} {Time} {Series} {Forecasting}},
\newblock in: \bibinfo{booktitle}{Proceedings of the 28th {ACM} {SIGKDD} {Conference} on {Knowledge} {Discovery} and {Data} {Mining}}, \bibinfo{year}{2022}, pp. \bibinfo{pages}{1567--1577}. \URLprefix \url{http://arxiv.org/abs/2206.09113}. \DOIprefix\doi{10.1145/3534678.3539396}, \bibinfo{note}{arXiv:2206.09113 [cs]}.
\bibitem[{Jin et~al.(2024)Jin, Wang, Ma, Chu, Zhang, Shi, Chen, Liang, Li, Pan, and Wen}]{jin_time-llm_2024}
\bibinfo{author}{M.~Jin}, \bibinfo{author}{S.~Wang}, \bibinfo{author}{L.~Ma}, \bibinfo{author}{Z.~Chu}, \bibinfo{author}{J.~Y. Zhang}, \bibinfo{author}{X.~Shi}, \bibinfo{author}{P.-Y. Chen}, \bibinfo{author}{Y.~Liang}, \bibinfo{author}{Y.-F. Li}, \bibinfo{author}{S.~Pan}, \bibinfo{author}{Q.~Wen}, \bibinfo{title}{Time-{LLM}: {Time} {Series} {Forecasting} by {Reprogramming} {Large} {Language} {Models}}, \bibinfo{year}{2024}. \URLprefix \url{http://arxiv.org/abs/2310.01728}. \DOIprefix\doi{10.48550/arXiv.2310.01728}, \bibinfo{note}{arXiv:2310.01728 [cs]}.
\bibitem[{Ansari et~al.(2024)Ansari, Stella, Turkmen, Zhang, Mercado, Shen, Shchur, Rangapuram, Arango, Kapoor, Zschiegner, Maddix, Wang, Mahoney, Torkkola, Wilson, Bohlke-Schneider, and Wang}]{ansari_chronos_2024}
\bibinfo{author}{A.~F. Ansari}, \bibinfo{author}{L.~Stella}, \bibinfo{author}{C.~Turkmen}, \bibinfo{author}{X.~Zhang}, \bibinfo{author}{P.~Mercado}, \bibinfo{author}{H.~Shen}, \bibinfo{author}{O.~Shchur}, \bibinfo{author}{S.~S. Rangapuram}, \bibinfo{author}{S.~P. Arango}, \bibinfo{author}{S.~Kapoor}, \bibinfo{author}{J.~Zschiegner}, \bibinfo{author}{D.~C. Maddix}, \bibinfo{author}{H.~Wang}, \bibinfo{author}{M.~W. Mahoney}, \bibinfo{author}{K.~Torkkola}, \bibinfo{author}{A.~G. Wilson}, \bibinfo{author}{M.~Bohlke-Schneider}, \bibinfo{author}{Y.~Wang}, \bibinfo{title}{Chronos: {Learning} the {Language} of {Time} {Series}}, \bibinfo{year}{2024}. \URLprefix \url{http://arxiv.org/abs/2403.07815}. \DOIprefix\doi{10.48550/arXiv.2403.07815}, \bibinfo{note}{arXiv:2403.07815 [cs]}.
\bibitem[{Zhou et~al.(2023)Zhou, Niu, Wang, Sun, and Jin}]{zhou_one_2023}
\bibinfo{author}{T.~Zhou}, \bibinfo{author}{P.~Niu}, \bibinfo{author}{X.~Wang}, \bibinfo{author}{L.~Sun}, \bibinfo{author}{R.~Jin}, \bibinfo{title}{One {Fits} {All}:{Power} {General} {Time} {Series} {Analysis} by {Pretrained} {LM}}, \bibinfo{year}{2023}. \URLprefix \url{http://arxiv.org/abs/2302.11939}. \DOIprefix\doi{10.48550/arXiv.2302.11939}, \bibinfo{note}{arXiv:2302.11939 [cs]}.
\bibitem[{Xue and Salim(2023)}]{Xue2023}
\bibinfo{author}{H.~Xue}, \bibinfo{author}{F.~D. Salim}, \bibinfo{title}{Promptcast: A new prompt-based learning paradigm for time series forecasting}, \bibinfo{year}{2023}. \URLprefix \url{https://arxiv.org/abs/2210.08964}. \href{http://arxiv.org/abs/2210.08964}{{\tt arXiv:2210.08964}}.
\bibitem[{Xu et~al.(2023)Xu, Xie, Qin, Tao, and Wang}]{xu2023parameterefficientfinetuningmethodspretrained}
\bibinfo{author}{L.~Xu}, \bibinfo{author}{H.~Xie}, \bibinfo{author}{S.-Z.~J. Qin}, \bibinfo{author}{X.~Tao}, \bibinfo{author}{F.~L. Wang}, \bibinfo{title}{Parameter-efficient fine-tuning methods for pretrained language models: A critical review and assessment}, \bibinfo{year}{2023}. \URLprefix \url{https://arxiv.org/abs/2312.12148}. \href{http://arxiv.org/abs/2312.12148}{{\tt arXiv:2312.12148}}.
\bibitem[{Han et~al.(2024)Han, Gao, Liu, Zhang, and Zhang}]{han2024parameterefficientfinetuninglargemodels}
\bibinfo{author}{Z.~Han}, \bibinfo{author}{C.~Gao}, \bibinfo{author}{J.~Liu}, \bibinfo{author}{J.~Zhang}, \bibinfo{author}{S.~Q. Zhang}, \bibinfo{title}{Parameter-efficient fine-tuning for large models: A comprehensive survey}, \bibinfo{year}{2024}. \URLprefix \url{https://arxiv.org/abs/2403.14608}. \href{http://arxiv.org/abs/2403.14608}{{\tt arXiv:2403.14608}}.
\bibitem[{Houlsby et~al.(2019)Houlsby, Giurgiu, Jastrzebski, Morrone, de~Laroussilhe, Gesmundo, Attariyan, and Gelly}]{houlsby2019parameterefficienttransferlearningnlp}
\bibinfo{author}{N.~Houlsby}, \bibinfo{author}{A.~Giurgiu}, \bibinfo{author}{S.~Jastrzebski}, \bibinfo{author}{B.~Morrone}, \bibinfo{author}{Q.~de~Laroussilhe}, \bibinfo{author}{A.~Gesmundo}, \bibinfo{author}{M.~Attariyan}, \bibinfo{author}{S.~Gelly}, \bibinfo{title}{Parameter-efficient transfer learning for nlp}, \bibinfo{year}{2019}. \URLprefix \url{https://arxiv.org/abs/1902.00751}. \href{http://arxiv.org/abs/1902.00751}{{\tt arXiv:1902.00751}}.
\bibitem[{Lester et~al.(2021)Lester, Al-Rfou, and Constant}]{lester2021powerscaleparameterefficientprompt}
\bibinfo{author}{B.~Lester}, \bibinfo{author}{R.~Al-Rfou}, \bibinfo{author}{N.~Constant}, \bibinfo{title}{The power of scale for parameter-efficient prompt tuning}, \bibinfo{year}{2021}. \URLprefix \url{https://arxiv.org/abs/2104.08691}. \href{http://arxiv.org/abs/2104.08691}{{\tt arXiv:2104.08691}}.
\bibitem[{Li and Liang(2021)}]{li2021prefixtuningoptimizingcontinuousprompts}
\bibinfo{author}{X.~L. Li}, \bibinfo{author}{P.~Liang}, \bibinfo{title}{Prefix-tuning: Optimizing continuous prompts for generation}, \bibinfo{year}{2021}. \URLprefix \url{https://arxiv.org/abs/2101.00190}. \href{http://arxiv.org/abs/2101.00190}{{\tt arXiv:2101.00190}}.
\bibitem[{Ba et~al.(2016)Ba, Kiros, and Hinton}]{ba2016layernormalization}
\bibinfo{author}{J.~L. Ba}, \bibinfo{author}{J.~R. Kiros}, \bibinfo{author}{G.~E. Hinton}, \bibinfo{title}{Layer normalization}, \bibinfo{year}{2016}. \URLprefix \url{https://arxiv.org/abs/1607.06450}. \href{http://arxiv.org/abs/1607.06450}{{\tt arXiv:1607.06450}}.
\bibitem[{Zhang and Sennrich(2019)}]{NEURIPS2019_1e8a1942}
\bibinfo{author}{B.~Zhang}, \bibinfo{author}{R.~Sennrich},
\newblock \bibinfo{title}{Root mean square layer normalization},
\newblock in: \bibinfo{editor}{H.~Wallach}, \bibinfo{editor}{H.~Larochelle}, \bibinfo{editor}{A.~Beygelzimer}, \bibinfo{editor}{F.~d\textquotesingle Alch\'{e}-Buc}, \bibinfo{editor}{E.~Fox}, \bibinfo{editor}{R.~Garnett} (Eds.), \bibinfo{booktitle}{Advances in Neural Information Processing Systems}, volume~\bibinfo{volume}{32}, \bibinfo{publisher}{Curran Associates, Inc.}, \bibinfo{year}{2019}. \URLprefix \url{https://proceedings.neurips.cc/paper_files/paper/2019/file/1e8a19426224ca89e83cef47f1e7f53b-Paper.pdf}.
\bibitem[{Su et~al.(2021)Su, Lu, Pan, Wen, and Liu}]{Su2021RoFormer}
\bibinfo{author}{J.~Su}, \bibinfo{author}{Y.~Lu}, \bibinfo{author}{S.~Pan}, \bibinfo{author}{B.~Wen}, \bibinfo{author}{Y.~Liu},
\newblock \bibinfo{title}{Roformer: Enhanced transformer with rotary position embedding},
\newblock \bibinfo{journal}{CoRR} \bibinfo{volume}{abs/2104.09864} (\bibinfo{year}{2021}). \URLprefix \url{https://arxiv.org/abs/2104.09864}. \href{http://arxiv.org/abs/2104.09864}{{\tt arXiv:2104.09864}}.
\bibitem[{Shazeer(2020)}]{Shazeer2020}
\bibinfo{author}{N.~Shazeer},
\newblock \bibinfo{title}{{GLU} variants improve transformer},
\newblock \bibinfo{journal}{CoRR} \bibinfo{volume}{abs/2002.05202} (\bibinfo{year}{2020}). \URLprefix \url{https://arxiv.org/abs/2002.05202}. \href{http://arxiv.org/abs/2002.05202}{{\tt arXiv:2002.05202}}.
\bibitem[{Aghajanyan et~al.(2020)Aghajanyan, Zettlemoyer, and Gupta}]{aghajanyan2020}
\bibinfo{author}{A.~Aghajanyan}, \bibinfo{author}{L.~Zettlemoyer}, \bibinfo{author}{S.~Gupta}, \bibinfo{title}{Intrinsic dimensionality explains the effectiveness of language model fine-tuning}, \bibinfo{year}{2020}. \URLprefix \url{https://arxiv.org/abs/2012.13255}. \href{http://arxiv.org/abs/2012.13255}{{\tt arXiv:2012.13255}}.
\bibitem[{Loshchilov and Hutter(2019)}]{adamw}
\bibinfo{author}{I.~Loshchilov}, \bibinfo{author}{F.~Hutter}, \bibinfo{title}{Decoupled weight decay regularization}, \bibinfo{year}{2019}. \URLprefix \url{https://arxiv.org/abs/1711.05101}. \href{http://arxiv.org/abs/1711.05101}{{\tt arXiv:1711.05101}}.
\bibitem[{Kudo and Richardson(2018)}]{kudo2018sentencepiecesimplelanguageindependent}
\bibinfo{author}{T.~Kudo}, \bibinfo{author}{J.~Richardson}, \bibinfo{title}{Sentencepiece: A simple and language independent subword tokenizer and detokenizer for neural text processing}, \bibinfo{year}{2018}. \URLprefix \url{https://arxiv.org/abs/1808.06226}. \href{http://arxiv.org/abs/1808.06226}{{\tt arXiv:1808.06226}}.
\bibitem[{Godahewa et~al.(2021)Godahewa, Bergmeir, Webb, Hyndman, and Montero-Manso}]{godahewa2021monash}
\bibinfo{author}{R.~Godahewa}, \bibinfo{author}{C.~Bergmeir}, \bibinfo{author}{G.~I. Webb}, \bibinfo{author}{R.~J. Hyndman}, \bibinfo{author}{P.~Montero-Manso},
\newblock \bibinfo{title}{Monash time series forecasting archive},
\newblock in: \bibinfo{booktitle}{Neural Information Processing Systems Track on Datasets and Benchmarks}, \bibinfo{year}{2021}.
\bibitem[{Zhou et~al.(2020)Zhou, Zhang, Peng, Zhang, Li, Xiong, and Zhang}]{Zhou2020Informer}
\bibinfo{author}{H.~Zhou}, \bibinfo{author}{S.~Zhang}, \bibinfo{author}{J.~Peng}, \bibinfo{author}{S.~Zhang}, \bibinfo{author}{J.~Li}, \bibinfo{author}{H.~Xiong}, \bibinfo{author}{W.~Zhang},
\newblock \bibinfo{title}{Informer: Beyond efficient transformer for long sequence time-series forecasting},
\newblock \bibinfo{journal}{CoRR} \bibinfo{volume}{abs/2012.07436} (\bibinfo{year}{2020}). \URLprefix \url{https://arxiv.org/abs/2012.07436}. \href{http://arxiv.org/abs/2012.07436}{{\tt arXiv:2012.07436}}.
\bibitem[{Shcherbakov et~al.(2013)Shcherbakov, Brebels, Shcherbakova, Tyukov, Janovsky, Kamaev et~al.}]{shcherbakov2013survey}
\bibinfo{author}{M.~V. Shcherbakov}, \bibinfo{author}{A.~Brebels}, \bibinfo{author}{N.~L. Shcherbakova}, \bibinfo{author}{A.~P. Tyukov}, \bibinfo{author}{T.~A. Janovsky}, \bibinfo{author}{V.~A. Kamaev}, et~al.,
\newblock \bibinfo{title}{A survey of forecast error measures},
\newblock \bibinfo{journal}{World applied sciences journal} \bibinfo{volume}{24} (\bibinfo{year}{2013}) \bibinfo{pages}{171--176}.
\bibitem[{Mart{\'i}nez et~al.(2019)Mart{\'i}nez, Fr{\'i}as, P{\'e}rez, and Rivera}]{Martinez2009}
\bibinfo{author}{F.~Mart{\'i}nez}, \bibinfo{author}{M.~P. Fr{\'i}as}, \bibinfo{author}{M.~D. P{\'e}rez}, \bibinfo{author}{A.~J. Rivera},
\newblock \bibinfo{title}{A methodology for applying k-nearest neighbor to time series forecasting},
\newblock \bibinfo{journal}{Artificial Intelligence Review} \bibinfo{volume}{52} (\bibinfo{year}{2019}) \bibinfo{pages}{2019--2037}. \URLprefix \url{https://doi.org/10.1007/s10462-017-9593-z}. \DOIprefix\doi{10.1007/s10462-017-9593-z}.

\end{thebibliography}

\end{document}